\documentclass[letterpaper]{article} 
\usepackage{aaai25}  
\usepackage{times}  
\usepackage{helvet}  
\usepackage{courier}  
\usepackage[hyphens]{url}  
\usepackage{graphicx} 
\usepackage{natbib}  
\usepackage{caption} 
\usepackage{algorithm}
\usepackage{algorithmic}
\usepackage[table]{xcolor}
\usepackage{newfloat}
\usepackage{listings}
\DeclareCaptionStyle{ruled}{labelfont=normalfont,labelsep=colon,strut=off} 
\floatstyle{ruled}
\newfloat{listing}{tb}{lst}{}
\floatname{listing}{Listing}
\usepackage{amssymb}
\usepackage{latexsym}
\usepackage{multirow}
\usepackage{pifont}
\usepackage{amsmath}
\usepackage{booktabs}
\usepackage{tabularx}
\usepackage{csquotes}
\usepackage[tikz]{mdframed}
\usetikzlibrary{shadows}
\usepackage{arydshln}
\usepackage{color, colortbl}
\usepackage{array}
\usepackage{enumitem}
\usepackage{bbding}
\newcommand{\eg}{\textit{e.g.,}\ }
\newcommand{\ie}{\textit{i.e.,}\ }
\newcommand{\itbf}[1]{\textit{\textbf{#1}}}
\newcommand{\tablerowcolor}{\rowcolor{gray!15}}
\usepackage{microtype}

\definecolor{question}{RGB}{223,223,223}
\definecolor{prompt}{RGB}{230,236,225}
\definecolor{context}{RGB}{236,215,192}
\definecolor{divide}{rgb}{1,0.,0}
\definecolor{conquer}{rgb}{0.25,0.25,0.90}
\definecolor{combine}{rgb}{0,0.8,0}
\definecolor{incorrect}{RGB}{182,36,35}
\definecolor{correct}{RGB}{16,222,45}
\definecolor{keyword}{RGB}{0,255,255}
\usepackage{inconsolata}
\usepackage{xcolor}
\definecolor{introred}{RGB}{248,203,173}

\usepackage{tikz}       %
\usepackage{tcolorbox}
\usepackage{makecell}
\usetikzlibrary{shapes.misc}
\usetikzlibrary{shadows}

\definecolor{backgroundcolor}{gray}{.9}

\newcommand{\green}[1]{\textcolor[RGB]{0,176,80}{#1}}
\newcommand{\red}[1]{\textcolor[RGB]{176,0,80}{#1}}

\title{Divide, Conquer and Combine: A Training-Free Framework for High-Resolution Image Perception in Multimodal Large Language Models}
\author {
    Wenbin Wang\textsuperscript{\rm 1}\thanks{Equal contribution.},
    Liang Ding\textsuperscript{\rm 2}$^{*}$,
    Minyan Zeng\textsuperscript{\rm 1},
    Xiabin Zhou\textsuperscript{\rm 3},
    Li Shen\textsuperscript{\rm 4},
    Yong Luo\textsuperscript{\rm 1}\thanks{Corresponding author.},
    Dacheng Tao\textsuperscript{\rm 5}
}
\affiliations {
    \textsuperscript{\rm 1}Wuhan University,
    \textsuperscript{\rm 2}The University of Sydney\\
    \textsuperscript{\rm 3}Jiangsu University,
    \textsuperscript{\rm 4}Sun Yat-Sen University,
    \textsuperscript{\rm 5}Nanyang Technological University\\
    wangwenbin97@whu.edu.cn, liangding.liam@gmail.com, luoyong@whu.edu.cn
}

\usepackage{bibentry}

\begin{document}

\maketitle

\begin{abstract}

Multimodal large language models (MLLMs) have experienced significant advancements recently, but still struggle to recognize and interpret intricate details in high-resolution (HR) images effectively. 
While state-of-the-art (SOTA) MLLMs claim to process images at 4K resolution, existing MLLM benchmarks only support up to 2K, leaving the capabilities of SOTA models on true HR images largely untested.
Furthermore, existing methods for enhancing HR image perception in MLLMs rely on computationally expensive visual instruction tuning.
To address these limitations, we introduce \itbf{HR-Bench}, the first deliberately designed benchmark to rigorously evaluate MLLM performance on \textbf{4K\&8K images}.
Through extensive experiments, we demonstrate that \textit{while downsampling HR images leads to vision information loss, leveraging complementary modalities, \eg text, can effectively compensate for this loss}. Building upon this insight, we propose \textbf{D}ivide, \textbf{C}onquer and \textbf{C}ombine (\textbf{DC$^2$}), a novel training-free framework for enhancing MLLM perception of HR images. 
\textbf{DC$^2$} follows a three-staged approach: 1) \textbf{Divide}: recursively partitioning the HR image into patches and merging similar patches to minimize computational overhead, 2) \textbf{Conquer}: leveraging the MLLM to generate accurate textual descriptions for each image patch, and 3) \textbf{Combine}: utilizing the generated text descriptions to enhance the MLLM's understanding of the overall HR image.
Extensive experiments show that: 1) the {\textit{SOTA MLLM achieves 63\% accuracy, which is markedly lower than the 87\% accuracy achieved by humans}} on \itbf{HR-Bench}; 2) our \textbf{DC$^2$} brings {\textit{consistent and significant improvements (a relative increase of +6\% on \itbf{HR-Bench} and +8\% on {\textbf{general multimodal} benchmarks}}}). The benchmark and code are released at \url{https://github.com/DreamMr/HR-Bench}.
\end{abstract}

\section{Introduction}
\begin{figure}[t]
  \begin{center}
  \includegraphics[width=1.\linewidth]{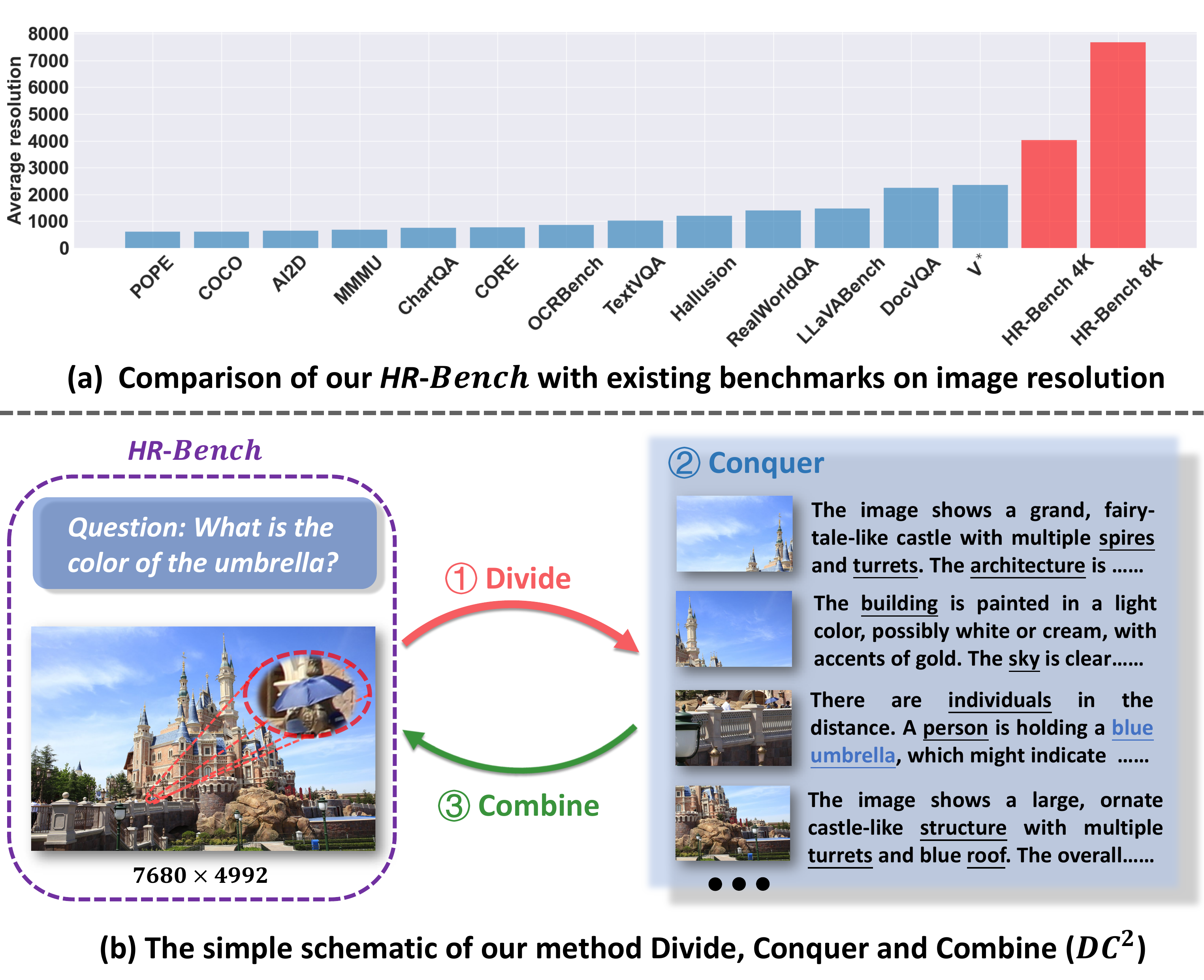}
  \end{center}
  \caption{An overview of our work. (a) We observe that the highest resolution in existing multimodal benchmarks is only 2K. To address the current lack of high-resolution (HR) multimodal benchmarks, we construct \itbf{HR-Bench} to evaluate the perception capabilities of MLLMs in HR images (up to 8K resolution). (b) We propose a training-free framework -- \textbf{D}ivide, \textbf{C}onquer and \textbf{C}ombine (\textbf{DC$^2$}), which recursively uses image patches to provide relevant text descriptions, helping existing MLLMs better perceive HR images.}
  \label{fig:motivation}
\end{figure}
Recent advancements in multimodal LLMs (MLLMs)~\cite{liu2024improved,internlmxcomposer2} have greatly enhanced their abilities in vision-language understanding, reasoning, and interaction~\cite{caffagni2024r}. This progress is primarily due to the integration of visual signals into Large Language Models (LLMs), allowing them to perceive the world visually. A key component of this process is the visual encoding strategy. However, most current MLLMs~\cite{liu2024improved,team2024chameleon} perceive images in a fixed resolution (\eg $336\times 336$). This simplification often results in significant shape distortion and blurring of high-resolution (HR) image content, which hurts the performance of MLLMs. Given that real-world images vary widely in resolution, this limitation poses substantial challenges for MLLMs across various applications.

To address this issue, recent studies improve MLLM's perceptual ability for HR image by carefully designing various strategies, which can be categorized into three types: 1) cropping-based methods~\cite{chen2024far,liu2024llavanext,li2024monkey}, 2) HR visual encoder~\cite{luo2024feast,ge2024convllava,lu2024deepseekvl}, and 3) visual search~\cite{wu2024v}. Although many advanced strategies have been proposed to enhance MLLM's perceptual ability for HR image, the current benchmark resolution is only up to 2K, as illustrated in Figure~\ref{fig:motivation} (a). Meanwhile, the most advanced MLLMs are now capable of handling 4K HR images~\cite{chen2024far,internlmxcomposer2_4khd}. This implies that \textbf{state-of-the-art (SOTA) MLLMs have not yet undergone rigorous validation on HR images}. Therefore, higher resolution benchmarks are needed in this field.

Firstly, to tackle the current lack of HR multimodal benchmarks, we introduce \itbf{HR-Bench}. This benchmark is designed to evaluate the ability of MLLMs to perceive HR images. \itbf{HR-Bench} is available in two versions: \itbf{HR-Bench 8K} and \itbf{HR-Bench 4K}. The \itbf{HR-Bench 8K} includes images with an average resolution of 8K, sourced from the open-source 8K resolution image dataset DIV8K~\cite{gu2019div8k} and Internet, with our manually annotated questions and answers. For \itbf{HR-Bench 4K}, we manually annotate the coordinates of objects relevant to the questions within the 8K image and crop these images to 4K resolution. This benchmark aims to systematically evaluate the ability of MLLM to perceive HR images, thus paving the way for future research.

Secondly, we conduct a series of experiments on the \itbf{HR-Bench} to explore the effects of image resolution on MLLMs. We select SOTA MLLMs~\cite{chen2024far,liu2024improved,Qwen-VL} to evaluate their performance across varying image resolutions (\eg 1K, 2K and 8K resolution). The experimental results indicate that downsampling HR images to a lower, fixed resolution leads to a significant \textit{loss of visual information}. This degradation increases the uncertainty in the model's output, making them more prone to errors. \textit{Notably, integrating information from other modalities (\eg text), proves effective in mitigating the adverse effects of lost visual information.}

Finally, we combine what we have learned above to design a new training-free framework which we call \textcolor{divide}{\ding{172}}\textbf{D}ivide, \textcolor{conquer}{\ding{173}}\textbf{C}onquer and \textcolor{combine}{\ding{174}}\textbf{C}ombine (\textbf{DC$^2$}). Our \textbf{DC$^2$} processes HR images by breaking them down into smaller, manageable image patches and using their accurate text descriptions to enhance MLLMs perception, as shown in Figure~\ref{fig:motivation} (b). Specifically, \textcolor{divide}{\ding{172}} we divide an HR image into smaller patches recursively until they match the resolution of the pretrained visual encoder (\eg $336\times 336$). To enhance computational efficiency, similar patches are merged. In the conquer stage\textcolor{conquer}{\ding{173}}, we use MLLM to generate text descriptions for each image patch. During the combine stage\textcolor{combine}{\ding{174}}, we aggregate the text descriptions and filter out hallucinations caused by the dividing stage. Directly using all text descriptions can hurt performance due to excessive input length. Inspired by~\citet{wu2024v}, we introduce a visual memory $\mathcal{M}$ to store objects which appear in the text description, and coordinates of image patches. In the inference stage, we use the user prompt to interact with $\mathcal{M}$, enabling MLLM to generate more precise text descriptions. Experiments demonstrate that our \textbf{DC$^2$} significantly improves performance on HR image benchmarks and outperforms existing methods on general multimodal benchmarks.
Our \textbf{contributions} are summarized as follows:
\begin{itemize}
  \item We introduce \itbf{HR-Bench} to systematically evaluate the perception ability of MLLMs in HR images. To the best of our knowledge, we are the first to propose an 8K image resolution benchmark for MLLMs.
  \item Based on our \itbf{HR-Bench}, we explore the impact of image resolution on MLLMs. we find that downsampling HR images reduces visual information, increasing uncertainty and errors in model outputs. Fortunately, adding proper textual information can effectively restore these lost information.
  \item Given our observation, we propose a training-free framework \textbf{DC$^2$} to effectively enhance the MLLM's perceive ability on HR images. Experimental results on our \itbf{HR-Bench} and general multimodal benchmarks using several advanced MLLMs, show that our approach brings consistent and significant improvements (up to +12.0\% accuracy). 
\end{itemize}

\section{Preliminaries and Related Work}
MLLMs generally include a \textbf{Visual Encoder}~\cite{radford2021learning} for extracting visual features and a \textbf{Large Language Model (LLM)}~\cite{touvron2023llama,touvron2023llama2,bai2023qwen,cai2024internlm2,glm2024chatglm} for decoding text sequences. Both the visual encoder and LLM are usually initialized from pre-trained models. The vision and language modalities can be connected by \textbf{Multimodal Connector} (\eg MLP). MLLMs generate sentences in an auto-regressive manner, predicting the probability distribution of the next token progressively. To maintain consistency with the image resolution used during visual encoder pre-training, MLLMs typically resize the image to a fixed resolution (\eg $336\times 336$ in LLaVA) before extracting visual features through the visual encoder. However, this simplification often results in significant shape distortion and blurring of HR image content. To address this issue, current solutions can be divided into \textbf{1) cropping-based methods}, \textbf{2) incorporating HR visual encoder methods}, and \textbf{3) visual search methods}.

\paragraph{Cropping-Based Methods.}
The representative cropping-based methods for HR MLLMs are introduced in LLaVA-NeXT~\cite{liu2024llavanext} and InternVL-v1.5~\cite{chen2024far}, which partition an image into several patches, each encoded separately by ViT~\cite{dosovitskiy2020image} and subsequently concatenated for LLM processing. Several methods have adopted cropping to scale up resolution~\cite{chen2024dragonfly,zhang2024beyond,liu2024infimm}.

\paragraph{HR Visual Encoder.}
Incorporating a HR visual encoder for HR image understanding does not substantially increase the number of visual tokens. Vary~\cite{wei2023vary} and Deepseek-VL~\cite{lu2024deepseekvl} harness SAM~\cite{kirillov2023segment} as a HR visual encoder to boost ViT's capabilities. Similarly, MiniGemini-HD~\cite{li2024mini}, LLaVA-HR~\cite{luo2024feast} and ConvLLaVA~\cite{ge2024convllava} leverage ConvNeXt~\cite{liu2022convnet} to handle HR images, utilizing cross-attention or adapter to extract visual features.

\begin{figure*}[thb]
  \begin{center}
  \includegraphics[width=1.\linewidth]{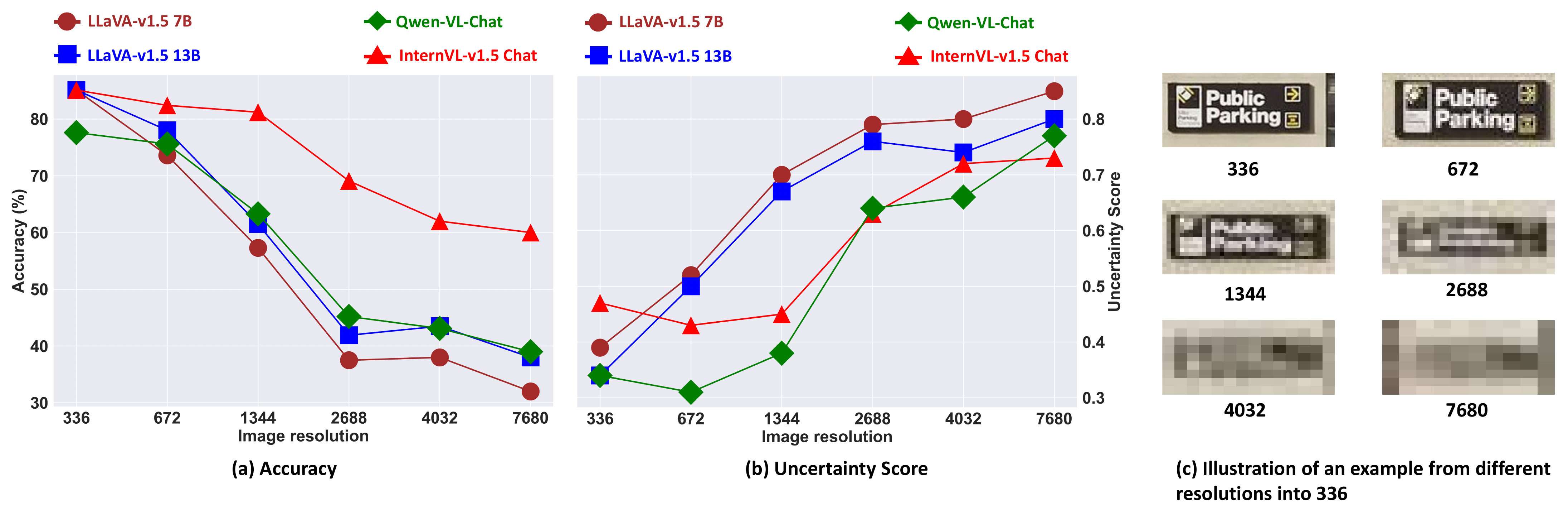}
  \end{center}
  \caption{Experimental results for accuracy and uncertainty scores under different image resolutions. We illustrate the accuracy (a) and uncertainty score (b) on four models with different image resolutions. Additionally, we visualize an example that is resized from different resolutions into 336 (c).}
  \label{fig:scale_resolution}
\end{figure*}

\paragraph{Visual Search.}
Inspired by key elements in the human visual search process,~\citet{wu2024v} introduce SEAL, a meta-architecture for MLLMs. SEAL is designed to actively reason about and seek out necessary visual information, a crucial capability for vision-intensive multimodal tasks, especially when dealing with HR images.

Despite numerous strategies proposed to enhance MLLMs' perceptual ability for HR images, the image resolution of current benchmarks~\cite{wu2024v,mathew2021docvqa,masry-etal-2022-chartqa,yue2023mmmu,Kembhavi2016ADI,li2023evaluating,MMBench,fu2023mme,yu2024mm,lu2024mathvista,ma-etal-2024-3am,han2023coremm,chen2024we,li2023seed2,liu2024hidden,lu2022learn} remains capped at 2K. In contrast, the latest MLLMs now handle 4K HR images~\cite{chen2024far,internlmxcomposer2_4khd}. This discrepancy indicates a pressing need for higher resolution MLLM benchmarks. Additionally, the factors influencing perceptual ability of MLLMs for HR images have not been thoroughly investigated.

\section{How does Image Resolution Affect MLLMs?}

\subsection{\itbf{HR-Bench}}
\label{sec:benchmark}
To systematically evaluate the effect of image resolution on MLLMs, we need a benchmark with sufficiently high resolution. We analyze the image resolution of 21 commonly used multimodal benchmarks and find that the benchmark with the highest image resolution is \textbf{V$^*$}, which is only $2246\times 1582$. The average resolution of the 21 multimodal benchmarks is $530\times 518$. However, it is known that the current SOTA MLLMs~\cite{chen2024far,internlmxcomposer2_4khd} are capable of handling 4K images. Thus, we introduce the \itbf{HR-Bench} using 200 8K resolution images from DIV8K~\cite{gu2019div8k} and the Internet. Our \itbf{HR-Bench} has significantly high resolution than other MLLM benchmarks -- 4$\times$ more than \textbf{V$^*$}.

\paragraph{\itbf{HR-Bench} Curation.}
\itbf{HR-Bench} consists two sub-tasks: \itbf{Fine-grained Single-instance Perception (FSP)} and \itbf{Fine-grained Cross-instance Perception (FCP)}. The \itbf{FSP} task includes 100 samples, challenging the MLLM to identify specific attributes such as color and material of an object. Similarity, the \itbf{FCP} task also comprises 100 samples but focuses on assessing the MLLM's ability to determine the relative positions between objects in an image. Both the images and questions are meticulously selected and crafted by human annotators to ensure it is challenging to ``guess'' the correct answer without accurately grounding the relevant objects in the image. In addition, the 8K images are cropped around the objects in question to produce 4K images. For clarity, the 8K resolution images are termed \itbf{HR-Bench 8K}, while the 4K resolution images are referred to as \itbf{HR-Bench 4K}. Examples of our benchmark can be found in the Appendix~D.

\paragraph{Evaluation of Protocol.}
To quantitatively compare MLLMs on our \itbf{HR-Bench}, we create multiple choice options for each question. Recognizing that MLLMs can be sensitive to the order of options in multiple choice questions, we use a more robust evaluation strategy called \textbf{Cyclic Permutation}~\cite{zheng2023large}. In particular, each question is presented to an MLLM $N$ times, with $N$ being the number of choices. Each time, the order of options are rotated to form a new prompt for the MLLMs. After completing $N$ passes, we calculate and report the average accuracy, ensuring a more reliable assessment.

\subsection{Pilot Experiments}
\label{sec:pilot_experient}
Despite the numerous advanced strategies proposed to handle HR images~\cite{chen2024far,internlmxcomposer2_4khd}, the impact of image resolution on MLLMs remains underexplored. Here, we raise two questions:
\begin{itemize}[label=\textendash]
    \item \itbf{How does image resolution affect MLLMs?}
    \item \itbf{How can we use answer to the above question to improve on prior methods?}
\end{itemize}

To answer these questions, we perform experiment on the existing SOTA MLLMs, selecting four widely used models: LLaVA-v1.5 7B \& 13B~\cite{liu2024improved}, Qwen-VL-Chat~\cite{Qwen-VL} and InternVL-v1.5-Chat~\cite{chen2024far}. These models cover various dimensions, including model scales, types of multimodal connectors, and types of visual encoders, enabling a more comprehensive analysis.

\paragraph{\textit{How does image resolution affect MLLMs?}}
We conduct experiments on our \itbf{HR-Bench}. We manually annotate the coordinates of relevant objects in each sample, and then crop the images centered on these coordinates to obtain images with different resolutions. During the cropping process, we maintain the original image aspect ratio. We use the following metrics to assess the impact of resolution for MLLMs: (1) \textbf{Accuracy} and (2) \textbf{Uncertainty Score}, which measures the model's confidence in generating the next token~\cite{zhouanalyzing}. A higher uncertainty score indicates greater uncertainty about the output, suggesting that the model's outputs are more likely to be inaccurate.

Figure~\ref{fig:scale_resolution} (a) and (b) illustrate that, across all models, \textbf{the accuracy significantly decreases and uncertainty score increases as the image resolution grows.} This can be intuitively explained by the significant loss of visual information during the downsampling of HR images. To prove this, Figure~\ref{fig:scale_resolution} (c) shows examples of images resized from various resolutions to 336. We observe that resizing from HR to low-resolution causes blurriness and results in a loss of detailed visual information.

\begin{mdframed}
[backgroundcolor=gray!40,shadow=true,roundcorner=8pt]
  \textbf{\textit{Finding 1: Downsampling higher-resolution images to a fixed resolution leads to greater visual information loss, increasing model output uncertainty, thereby causing output errors.}}
\end{mdframed}

\begin{figure}[thb]
  \begin{center}
  \includegraphics[width=1.0\linewidth]{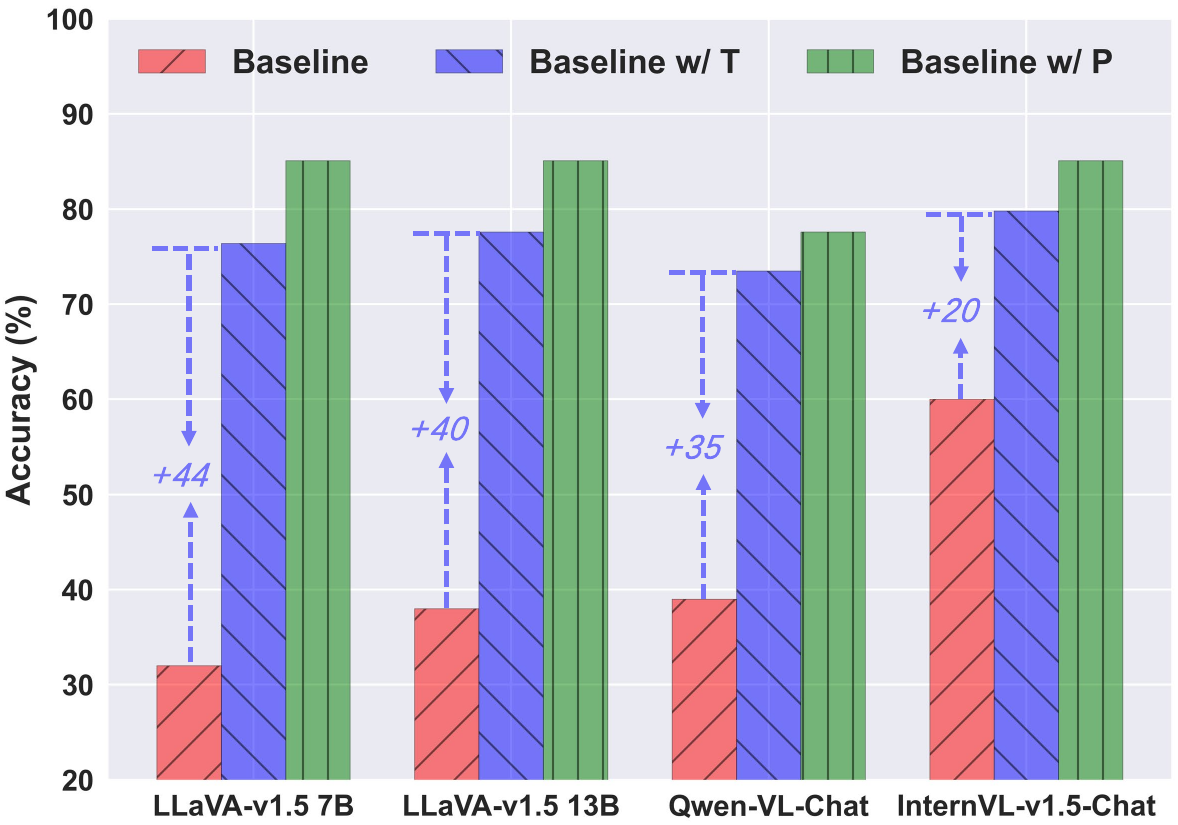}
  \end{center}
  \caption{The effect of incorporating rich text description on model performance.  
  ``\textbf{T}'' represents text descriptions. ``\textbf{P}'' represents key image regions.
  }
  \label{fig:with_content}
  \vspace{-0.5cm}
\end{figure}

\paragraph{\textit{Can we use language modality information to compensate for the missing visual information?}}
To answer this question, we design two experiments: 1) We manually provide rich text descriptions of the images in our \itbf{HR-Bench 8K}, detailing the attributes of the objects (\eg color) and the relative positions between them in the image (``\textbf{T}''). We \textbf{DO NOT} directly provide the answer to the question. 2) We extract the key image region, which the MLLM can rely on to generate correct answers, and replace the HR input to prevent visual information loss during downsampling (``\textbf{P}''). As shown in Figure~\ref{fig:with_content}, we find that 1) by introducing rich text descriptions, the performance is significantly improved on our \itbf{HR-Bench 8K}; and 2) incorporating text descriptions can achieve performance comparable to preserving key regions of the image.

\begin{figure*}[t]
  \begin{center}
  \includegraphics[width=1.0\linewidth]{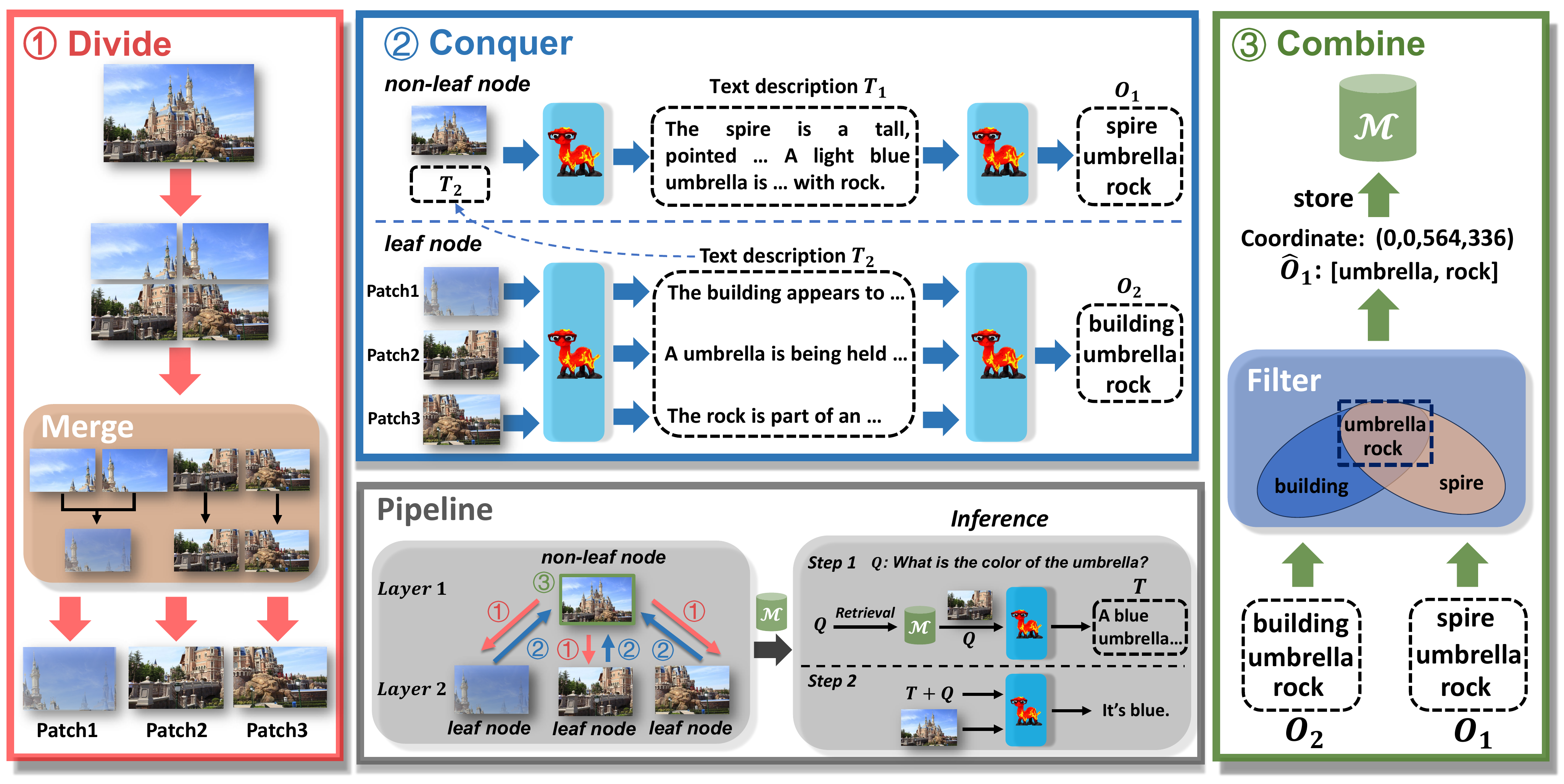}
  \end{center}
  \caption{Detailed illustration of our proposed schema \textbf{DC$^2$} with a running example. \textcolor{divide}{\ding{172}} We divide the image into four image patches and then merge the patches that have a high degree of similarity. \textcolor{conquer}{\ding{173}} We use MLLMs to generate text descriptions and object information from image. \textcolor{combine}{\ding{174}} We filter out uncertainty objects and then store the coordinates of the actually existing objects.}
  \label{fig:framework}
\end{figure*}

\begin{mdframed}[backgroundcolor=gray!40,shadow=true,roundcorner=8pt]
  \textbf{\textit{Finding 2: The visual information loss due to downsampling in HR images can be compensated by relevant textual information.}}
\end{mdframed}

\section{Methodology}
\paragraph{Overviews.}
Based on the aforementioned findings, we propose a novel training-free framework --- \textbf{\textcolor{divide}{\ding{172}} D}ivide, \textbf{\textcolor{conquer}{\ding{173}} C}onquer and \textbf{\textcolor{combine}{\ding{174}} C}ombine (\textbf{DC$^2$}) (see Figure~\ref{fig:framework}). \textit{The design principle of our method is to use the accurate text descriptions of image patches to help MLLM better perceive HR image.} To achieve this, we first recursively split an image into image patches until they reach the resolution defined by the pretrained vision encoder (\eg $336\times 336$), merging similar patches for efficiency (\textbf{Divide}). Next, we utilize MLLM to generate text description for each image patch and extract objects mentioned in the text descriptions (\textbf{Conquer}). Finally, we filter out hallucinated objects resulting from image division and store the coordinates of the image patches which objects appear (\textbf{Combine}). During the inference stage, we retrieve the related image patches according to the user prompt to provide accurate text descriptions.

\subsection{\textcolor{divide}{\ding{172}} Dividing: Image Division}
The goal of the \textbf{Divide} stage is to decompose the image into the resolution defined by pretrained visual encoder (\eg $336\times 336$), avoiding excessive visual information loss due to downsampling. However, we find that decomposing HR images into an excessive number of image patches disrupts object integrity, hindering the acquisition of global image information. Inspired by CNN~\cite{lecun1989backpropagation,he2016deep}, we recursively decompose the image, dividing it into four equal parts until the resolution defined by pretrained vision encoder is reached, thereby reducing the loss of global information. As shown in Figure~\ref{fig:framework}, the entire process can be visualized as a tree-like structure.

Specifically, given an image $v_l$, we crop $v_l$ into four patches. Mathematically, this operation can be described as:
\begin{equation}
 \{\overline{v}^i\}_{i=1}^{4} = \mathcal{F}_{crop}(v_l),
\end{equation}
where $l$ represents the indices of the current recursive layer and $i$ represents the $i-$th image patch. $\mathcal{F}_{crop}(\cdot)$ is the cropping function used to split the image into four image patches.

However, it is not efficient to perform recursion for each image patch. In fact, visual signals have high redundancy~\cite{bolyatoken}. To optimize computational efficiency, we merge (\ie by averaging) the image patches with similarity greater than $\theta$ by performing hierarchical clustering ($HC$) on the image patches.  This process can be formulated as follows:
\begin{equation}
    [C_1,...,C_k] = HC(\{\overline{v}^i\}_{i=1}^{4},\theta),
\end{equation}
\begin{equation}
    v_{l+1}^{i} = \frac{1}{|C_i|}\sum_{\overline{v}\in C_i} \overline{v},
\end{equation}
where the $k$ represents the number of clusters and $C_i$ represents the $i-$th cluster. Thus, the output of \textbf{Divide} stage is $\{v_{l+1}^{i}\}_{i=1}^k$ in current recursive layer. Then, the $v_{l+1}^i$ serves as the input to the next recursive layer.

\subsection{\textcolor{conquer}{\ding{173}} Conquering: Local Image Perception}
In the \textbf{Conquer} stage, image patches obtained in dividing process are used to generate text description and extract objects information by the MLLM. Specifically, given an image $v_l$, we firstly utilize MLLM to generate text description $T_l$. Then, we identify the main objects $O_l$ mentioned in the generated text description $T_l$. We denote the image patch which does not branch out to any other image patches as leaf node, while others are called non-leaf nodes. For a leaf node, we directly use MLLM to generate text description $T_l$. For non-leaf node, we concatenate the text descriptions from image patches $\{v_{l+1}^{i}\}_{i=1}^k$ (\ie $T_{l+1}$) to generate text description for image $v_l$. This is formulated as follows:
\begin{equation}
    \begin{cases}
    T_l, O_l = \mathcal{F}_{leaf}(v_l) &  \text{if } v_l \text{ is a leaf node} \\
    T_l, O_l = \mathcal{F}_{non-leaf}(v_l,T_{l+1}) & \text{otherwise}
\end{cases},   
\end{equation}
where the $\mathcal{F}_{leaf}(\cdot)$ is used to generate the text description $T_l$ and extract objects $O_l$ for leaf nodes while $\mathcal{F}_{non-leaf}(\cdot)$ is used for non-leaf nodes. The implementation of $\mathcal{F}_{leaf}(\cdot)$ and $\mathcal{F}_{non-leaf}(\cdot)$ can be found in the Appendix B.

\subsection{\textcolor{combine}{\ding{174}} Combining: Global Fusion}
In the \textbf{Combine} stage, we aggregate the information from image patches. Actually, image division disrupts the integrity of objects leading to output with object hallucination. Therefore, we need to \textbf{filter} out object hallucination caused by image division. Additionally, using text descriptions of all image patches can result in excessively long input text, which hurt performance during inference. Inspired by~\citet{wu2024v}, we introduce visual memory $\mathcal{M}$. We obtain the coordinates of the image patch where each object is located and \textbf{store} them in visual memory $\mathcal{M}$. During inference, we retrieve the image patches containing the objects mentioned in the user prompt and generate text descriptions. We use $(x,y,w,h)$ to represent the coordinate of image patch (\ie bounding box). The $x$ and $y$ represent the coordinates of the left and top in the global image. The $w$ and $h$ represent the width and height of the image patch respectively.

\paragraph{Filter.} One direct approach is calculating the uncertainty~\cite{zhouanalyzing} by the probability of autoregressive decoding for each object. However, this method tends to be inefficient for filtering out hallucinated objects. Indeed, for a real existing object, it will be found by the MLLM in successive recursive layers. Based on this, we take the intersection of $O_l$ and $O_{l+1}$, considering the objects $\widehat{O}_l$ in both sets to be actually existing. This can be formulated as follows:
\begin{equation}
    \widehat{O}_l = O_l \cap O_{l+1}.
\end{equation}

\paragraph{Storing in the Visual Memory $\mathcal{M}$.} After obtaining the actually existing objects $\widehat{O}_l$ and the coordinates of image patches, we store them in the visual memory $\mathcal{M}$. Two issues arise during storage: 1) overlapping image patches for the same object, and 2) coordinate representation of merged image patches. To address overlapping image patches, we apply Non-Maximum Suppression (NMS) to retain the patch that best represents the object. For coordinate representation, we save the coordinates before merging the image patches.

\subsection{Inference Details}
In the inference stage, we utilize the user prompt to interact with visual memory $\mathcal{M}$. Specifically, given a user prompt $Q$, we use a textual retriever~\cite{izacard2022unsupervised} to retrieve related objects with confidence levels exceed $\alpha$. Subsequently, we obtain the image patches for the retrieved objects to allow MLLM to generate accurate text descriptions $T$. Finally, we concatenate the accurate text descriptions $T$ with user prompt $Q$ and utilize the MLLM to generate the final response.

\section{Experiments}

\subsection{Evaluation on \itbf{HR-Bench 8K}}
\subsubsection{Overall performance.}
As shown in Table~\ref{table:hrbench}, the most proficient open-source MLLM, InternVL2-llama3-76B~\cite{chen2023internvl}, achieves accuracy of 61.4\% on \itbf{HR-Bench 8K}. Even the most advanced models, Gemini 1.5 Flash~\cite{reid2024gemini}, GPT4o~\cite{achiam2023gpt}, QWen-VL-max~\cite{Qwen-VL} achieve accuracies of 62.8\%, 55.5\% and 52.5\% on \itbf{HR-Bench 8K}. The results demonstrate that existing MLLMs still have a significant gap compared to humans in their perception of HR images. For more experimental analysis, see the appendix C.

\begin{table}[thb]
  \centering
  \setlength{\tabcolsep}{3mm}
  \begin{tabular}{lccc}
  \toprule
                 \multirow{2}{*}{\textbf{Method}}     &  \multicolumn{3}{c}{\itbf{HR-Bench 8K}} \\ \cmidrule(lr){2-4}
                          & \textbf{\textit{FSP}$\uparrow$}    & \textbf{\textit{FCP}$\uparrow$}    & \textbf{\textit{Avg.}$\uparrow$}    \\ \hline
  \textbf{Human}                  &   \textbf{94.0}     &   \textbf{79.5}     &   \textbf{86.8}     \\
  Random Guess           &   25.0     &  25.0      &   25.0     \\ \hline
  \multicolumn{4}{c}{\textbf{\textit{Open-source MLLMs}}} \\ \hline
  \textbf{InternVL-2-llama3-76B} & 69.0  & \textbf{53.8}  & \textbf{61.4}  \\
  InternVL-1.5-26B         & \textbf{69.3}  & 46.5   & 57.9  \\
  Xcomposer2-4kHD-7B   &  55.3  & 47.3   & 51.3  \\
  LLaVA-1.6-34B         & 44.5   & 50.3  & 47.4  \\
  LLaVA-HR-X-13B      & 49.5   & 44.3  & 46.9  \\
  LLaVA-HR-X-7B        & 42.0     & 41.3  & 41.6  \\
  CogVLM-Chat-17B            & 42.5   & 39.8  & 41.1  \\
  LLaVA-1.6-7B           & 37.2   & 44.2   & 40.8   \\
  Phi3-Vision-4.2B          & 43.3  & 37.8  & 40.5   \\
  Yi-VL-34B            & 39.5   & 38.5   & 39.0     \\
  \hline
  \multicolumn{4}{c}{\textbf{\textit{Commercial chatbot systems}}} \\ \hline
  \textbf{Gemini 1.5 Flash}            & \textbf{69.2}  & \textbf{56.7}   & \textbf{62.8}  \\
  GPT4o                  & 62.0     & 49.0     & 55.5   \\
  QWen-VL-max                   & 54.0     & 51.0     & 52.5   \\
  \hline
  \multicolumn{4}{c}{\textbf{\textit{w/ our DC$^2$}}} \\ \hline
  \textbf{InternVL-1.5 26B}      &   \textbf{75.0}  &  \textbf{47.5}  &  \textbf{61.3} \\
  \tablerowcolor \quad $\Delta (\uparrow)$ & +5.7 & +1.0  & +3.4   \\ \hdashline
  LLaVA-v1.6 7B        &   40.5  & 45.0   & 42.3  \\
  \tablerowcolor \quad $\Delta (\uparrow)$ & +3.3 & +0.8  & +2.1   \\ \hdashline
  LLaVA-v1.5 13B        &  40.0   & 41.0   & 40.5  \\
  \tablerowcolor \quad $\Delta (\uparrow)$ & +2.5 & +3.0  & +2.7   \\ \hdashline
  Yi-VL 6B        &  39.0   & 41.0   &  40.0 \\
  \tablerowcolor \quad $\Delta (\uparrow)$ & +0.5 & +1.7  & +1.1   \\
   \bottomrule
  \end{tabular}
  \caption{Results of different models on \itbf{HR-Bench 8K}. The best performance in each task is in-bold. The ``$\Delta (\uparrow)$'' represents the performance gains of our \textbf{DC$^2$} against the baselines. Due to space limitations, only the results of the top 10 open-source MLLMs and the top 3 commercial chatbot systems are presented here.}
  \label{table:hrbench}
  \end{table}

\subsubsection{Our DC$^2$ brings consistent improvements on \itbf{HR-Bench 8K}.}
We observe that our \textbf{DC$^2$} achieves consistent and significant improvement across four models and two sub-tasks. Our \textbf{DC$^2$} brings a maximum of 5.7\% and 3.0\% accuracy improvement on \itbf{FSP} and \itbf{FCP} respectively. Additionally, InternVL-v1.5 with our \textbf{DC$^2$} surpasses the current SOTA Gemini 1.5 Flash in the \itbf{FSP} sub-task, achieving an accuracy of 75.0\%. The results show that our method has a clear advantage with HR images.

\subsection{General Multimodal Benchmarks Evaluation}

To verify that our \textbf{DC$^2$} is not only applicable to HR images, we also conduct experiments on general multimodal benchmarks. As shown in Table~\ref{table:general_bench}, our \textbf{DC$^2$} not only brings up to a 12\% improvement in accuracy on 2K resolution MLLM benchmark \itbf{V$^*$}~\cite{wu2024v} but also shows significant improvements in object hallucination evaluation POPE~\cite{li2023evaluating} and comprehensive multimodal benchmark MME~\cite{fu2023mme}.
\begin{table}[hbt]
\centering
  \setlength{\tabcolsep}{3.5mm}
  \begin{tabular}{llcl}
  \toprule
  \textbf{Method}        & \itbf{V$^*\uparrow$} & \itbf{POPE$\uparrow$} & \itbf{MME$\uparrow$} \\ \hline
  Yi-VL-6B &  40.9  &   83.1   &  1902.7   \\ 
  \tablerowcolor \quad \textbf{+DC$^2$}        &   \textbf{46.2}      &    \textbf{83.2}      &    \textbf{1918.2}   \\ \hline 
  LLaVA-v1.5-7B  & 46.2   &  85.6    & 1755.9    \\ 
 \tablerowcolor \quad \textbf{+DC$^2$}        &   \textbf{57.3}      &   \textbf{86.8}       &   \textbf{1778.7}    \\ \hline
  LLaVA-v1.5 13B  & 42.7   &  85.5    & 1773.6    \\ 
  \tablerowcolor \quad \textbf{+DC$^2$}        &    \textbf{54.7}     &   \textbf{86.5}       &  \textbf{1779.1}      \\
  \bottomrule
  \end{tabular}
  \caption{Evaluation on broader range of general multimoadal benchmarks.  \textbf{DC$^2$} can also bring significant improvements on general multimodal benchmarks. The results are measured by VLMEVALKIT~\cite{duan2024vlmevalkit}.}
  \label{table:general_bench}
  \end{table}

\begin{figure*}[hbpt]
    \begin{center}
    \includegraphics[width=0.95\linewidth]{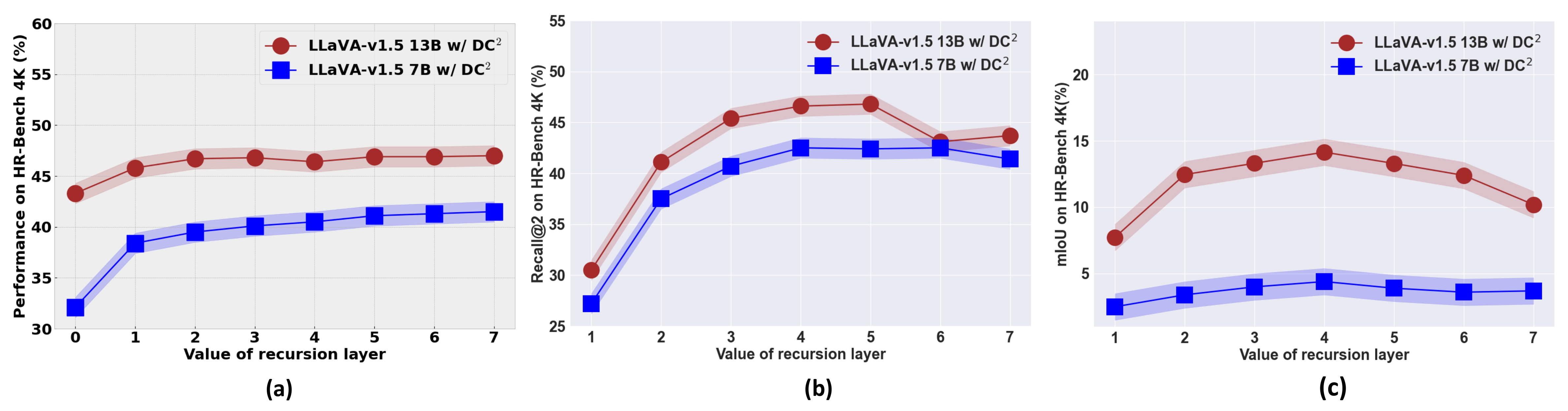}
    \end{center}
    \caption{
    Effect of recursion layers on \itbf{HR-Bench 4K}. (a) Overall performance, (b) Recall@2, (c) mIoU scores.
    }
    \label{fig:when_why}
\end{figure*}

\subsection{Ablation Study}
In Table~\ref{table:ablation}, we explore different modules, including merge, retrieval, filter, visual memory and recursive crop. The merge module causes a minor 0.5\% performance drop but enhances inference efficiency by reducing image patches. Excluding visual memory necessitates generating text descriptions for image patches during inference, leading to longer inputs and a 2.6\% performance drop. Removing the filter compromises object integrity in images, causing incorrect text descriptions and a 4.6\% performance decrease. Omitting recursive cropping severely impacts object integrity and increases input text length, resulting in a substantial 10.2\% performance decline.

\begin{table}[ht]
\centering
\begin{tabular}{lccl}
\toprule
\textbf{Method} &  \itbf{V$^*$}$\uparrow$  & \itbf{HR-Bench 8K}$\uparrow$  & \itbf{Avg.}$\uparrow$ \\ \hline
\textbf{DC$^2$} &  57.3  &  \textbf{39.5}  & 48.4 \\ \hdashline
\quad \textit{w/o merge} & \textbf{58.3}  & 39.5 & \textbf{48.9}  \\
\quad \textit{w/o visual memory}  & 55.6   & 36.0 & 45.8 \\
\quad \textit{w/o filter}   &  52.6  &  35.0 & 43.8 \\
\quad \textit{w/o recursive crop}   &  43.9  & 32.5   & 38.2 \\
\bottomrule
\end{tabular}
\caption{Ablation studies of our \textbf{DC$^2$}. We conduct experiments on \itbf{V$^*$} and \itbf{HR-Bench 8K} using LLaVA-v1.5 7B.}
\label{table:ablation}
\vspace{-0.3cm}
\end{table}

\subsection{Trade-off Between Performance and Efficiency}
\label{sec:efficiency} 
Researchers may have concerns regarding the efficiency of \textbf{DC$^2$}. To address this, we present Table~\ref{table:efficiency}, which illustrates the relationship between throughput and accuracy under various $\theta$ values (used to merge image patches), comparing these results with the SOTA method visual search~\cite{wu2024v}. As depicted, a decrease in $\theta$ leads to a continuous increase in accuracy on the \itbf{HR-Bench 8K}, accompanied by a decrease in efficiency. Notably, \textbf{DC$^2$} achieves higher accuracy than visual search for equivalent throughput, indicating a reasonable trade-off between performance and efficiency. 
\begin{table}[htb]
\centering
\setlength{\tabcolsep}{5.5mm}
\begin{tabular}{lcc}
\toprule
\textbf{Method}   & \itbf{Throughput}$\uparrow$ & \itbf{Acc.}$\uparrow$ \\ \hline
\textbf{DC}$^2$~\textit{w/o merge} &    2.8        &   \textbf{39.5}   \\
\textbf{DC}$^2$($\theta=0.1$) &     3.1       &   \textbf{39.5}  \\
\textbf{DC}$^2$($\theta=0.2$) &      4.6      &   36.5   \\
\textbf{DC}$^2$($\theta=0.3$) &      \textbf{5.0}      &   35.5   \\ \hdashline
Visual Search   &  4.6   & 35.6  \\
\bottomrule
\end{tabular}
\caption{Performance and inference efficiency. We illustrate the correlation between throughput (samples per minute) and the accuracy of the LLaVA-v1.5 7B enhanced with the proposed \textbf{DC}$^2$ across varying $\theta$ values on the \itbf{HR-Bench 8K}. Additionally, we also compare with SOTA method Visual Search, which is also used for HR images.}
\label{table:efficiency}
\vspace{-0.3cm}
\end{table}

\subsection{When and Why Does Our Method Work?}

Reviewing the design principles of \textbf{DC$^2$}: using text descriptions of image patches to help MLLM better perceive HR image. 
To explore the underlying mechanism of \textbf{DC$^2$}, we perform experiments that help address the following questions:

\itbf{1. Does increasing the number of image patches improve performance?}
We illustrate the relationship between the number of recursion layers and accuracy on \itbf{HR-Bench 4K} using LLaVA-v1.5 7B \& 13B. Figure~\ref{fig:when_why} (a) shows that 1) increasing the number of layers significantly improves accuracy; 2) however, as the recursion layers increase, the performance improvement gradually slows down. We observe that as the recursion layers increase, the performance of \itbf{FSP} improves significantly, but \itbf{FCP} appears to even slightly decline.
\textit{\textbf{More image patches reduce visual information loss, benefiting the \itbf{FSP} task.}}

\itbf{2. Can DC$^2$ provide precise text descriptions to compensate for the absence of visual information?} To demonstrate that our \textbf{DC$^2$} can provide precise information about objects, we employ the widely used evaluation metric Recall@2 to assess the performance of retrieving pertinent objects from visual memory $\mathcal{M}$. Additionally, we utilize mIoU to provide a more precise quantification of the overlap between the predicted bounding boxes derived from our \textbf{DC$^2$} and ground truth. As shown in Figure~\ref{fig:when_why}, the results show that 1) within a reasonable range of recursion layers, increasing the depth of recursion layers can yield more accurate object location information. 2) The more accurate the object location information provided by MLLM, the higher the accuracy on \itbf{HR-Bench 4K}.
\textit{\textbf{DC$^2$ can determine the position of objects in the image, thereby providing more accurate text description to compensate for missing visual information.}}

\subsection*{\ding{43} A Note on More Details in the Appendix}
See Appendix~A for \itbf{HR-Bench} details, Appendix~B for \textbf{DC$^2$} implementation, Appendix~C for additional experiment results, and Appendix~D for case studies.

\section{Conclusion}
In this paper, we propose an 8K image resolution benchmark, namely \itbf{HR-Bench} and introduce a training-free framework --- \textbf{D}ivide, \textbf{C}onquer and \textbf{C}ombine (\textbf{DC$^2$}). We systematically evaluate 28 open-source and commercial models on \itbf{HR-Bench}. From the results, we mainly conclude that: (1) MLLMs currently fall significantly short of humans in perceiving HR images; (2) current MLLMs lose a significant amount of visual information when resizing HR images to low-resolution, but this loss can be compensated for with text information; (3) our \textbf{DC$^2$} improves the current MLLMs' ability to perceive HR images. In the future, we will explore advanced token compression technologies, such as token merging-based methods for more efficient processing of images at any resolution, which could further enhance the MLLM's ability for high-resolution perception.

\bibliography{main}

\begin{thebibliography}{67}
\providecommand{\natexlab}[1]{#1}

\bibitem[{Abdin et~al.(2024)Abdin, Jacobs, Awan, Aneja, Awadallah, Awadalla,
  Bach, Bahree, Bakhtiari, Behl et~al.}]{abdin2024phi}
Abdin, M.; Jacobs, S.~A.; Awan, A.~A.; Aneja, J.; Awadallah, A.; Awadalla, H.;
  Bach, N.; Bahree, A.; Bakhtiari, A.; Behl, H.; et~al. 2024.
\newblock Phi-3 technical report: A highly capable language model locally on
  your phone.
\newblock \emph{arXiv preprint}.

\bibitem[{Achiam et~al.(2023)Achiam, Adler, Agarwal, Ahmad, Akkaya, Aleman,
  Almeida, Altenschmidt, Altman, Anadkat et~al.}]{achiam2023gpt}
Achiam, J.; Adler, S.; Agarwal, S.; Ahmad, L.; Akkaya, I.; Aleman, F.~L.;
  Almeida, D.; Altenschmidt, J.; Altman, S.; Anadkat, S.; et~al. 2023.
\newblock Gpt-4 technical report.
\newblock \emph{arXiv preprint}.

\bibitem[{Bai et~al.(2023{\natexlab{a}})Bai, Bai, Chu, Cui, Dang, Deng, Fan,
  Ge, Han, Huang et~al.}]{bai2023qwen}
Bai, J.; Bai, S.; Chu, Y.; Cui, Z.; Dang, K.; Deng, X.; Fan, Y.; Ge, W.; Han,
  Y.; Huang, F.; et~al. 2023{\natexlab{a}}.
\newblock Qwen technical report.
\newblock \emph{arXiv preprint}.

\bibitem[{Bai et~al.(2023{\natexlab{b}})Bai, Bai, Yang, Wang, Tan, Wang, Lin,
  Zhou, and Zhou}]{Qwen-VL}
Bai, J.; Bai, S.; Yang, S.; Wang, S.; Tan, S.; Wang, P.; Lin, J.; Zhou, C.; and
  Zhou, J. 2023{\natexlab{b}}.
\newblock Qwen-VL: A Versatile Vision-Language Model for Understanding,
  Localization, Text Reading, and Beyond.
\newblock \emph{arXiv preprint}.

\bibitem[{Bavishi et~al.(2023)Bavishi, Elsen, Hawthorne, Nye, Odena, Somani,
  and Ta\c{s}\i{}rlar}]{fuyu-8b}
Bavishi, R.; Elsen, E.; Hawthorne, C.; Nye, M.; Odena, A.; Somani, A.; and
  Ta\c{s}\i{}rlar, S. 2023.
\newblock Introducing our Multimodal Models.

\bibitem[{Bolya et~al.(2023)Bolya, Fu, Dai, Zhang, Feichtenhofer, and
  Hoffman}]{bolyatoken}
Bolya, D.; Fu, C.-Y.; Dai, X.; Zhang, P.; Feichtenhofer, C.; and Hoffman, J.
  2023.
\newblock Token Merging: Your {ViT} but Faster.
\newblock In \emph{ICLR}.

\bibitem[{Caffagni et~al.(2024)Caffagni, Cocchi, Barsellotti, Moratelli, Sarto,
  Baraldi, Cornia, and Cucchiara}]{caffagni2024r}
Caffagni, D.; Cocchi, F.; Barsellotti, L.; Moratelli, N.; Sarto, S.; Baraldi,
  L.; Cornia, M.; and Cucchiara, R. 2024.
\newblock The (r) evolution of multimodal large language models: A survey.
\newblock \emph{arXiv preprint}.

\bibitem[{Cai et~al.(2024)Cai, Cao, Chen, Chen, Chen, Chen, Chen, Chen, Chen,
  Chu, Dong, Duan, Fan, Fei, Gao, Ge, Gu, Gu, Gui, Guo, Guo, He, Hu, Huang,
  Jiang, Jiao, Jin, Lei, Li, Li, Li, Li, Li, Li, Liu, Liu, Hong, Liu, Liu, Liu,
  Lv, Lv, Lv, Ma, Ma, Ma, Ning, Ouyang, Qiu, Qu, Shang, Shao, Song, Song, Sui,
  Sun, Sun, Tang, Wang, Wang, Wang, Wang, Wang, Wang, Wang, Wei, Weng, Wu,
  Xiong, Xu, Xu, Yan, Yan, Yang, Ye, Ying, Yu, Yu, Zang, Zhang, Zhang, Zhang,
  Zhang, Zhang, Zhang, Zhang, Zhang, Zhang, Zhang, Zhang, Zhao, Zhao, Zhao,
  Zhou, Zhou, Zhuo, Zou, Qiu, Qiao, and Lin}]{cai2024internlm2}
Cai, Z.; Cao, M.; Chen, H.; Chen, K.; Chen, K.; Chen, X.; Chen, X.; Chen, Z.;
  Chen, Z.; Chu, P.; Dong, X.; Duan, H.; Fan, Q.; Fei, Z.; Gao, Y.; Ge, J.; Gu,
  C.; Gu, Y.; Gui, T.; Guo, A.; Guo, Q.; He, C.; Hu, Y.; Huang, T.; Jiang, T.;
  Jiao, P.; Jin, Z.; Lei, Z.; Li, J.; Li, J.; Li, L.; Li, S.; Li, W.; Li, Y.;
  Liu, H.; Liu, J.; Hong, J.; Liu, K.; Liu, K.; Liu, X.; Lv, C.; Lv, H.; Lv,
  K.; Ma, L.; Ma, R.; Ma, Z.; Ning, W.; Ouyang, L.; Qiu, J.; Qu, Y.; Shang, F.;
  Shao, Y.; Song, D.; Song, Z.; Sui, Z.; Sun, P.; Sun, Y.; Tang, H.; Wang, B.;
  Wang, G.; Wang, J.; Wang, J.; Wang, R.; Wang, Y.; Wang, Z.; Wei, X.; Weng,
  Q.; Wu, F.; Xiong, Y.; Xu, C.; Xu, R.; Yan, H.; Yan, Y.; Yang, X.; Ye, H.;
  Ying, H.; Yu, J.; Yu, J.; Zang, Y.; Zhang, C.; Zhang, L.; Zhang, P.; Zhang,
  P.; Zhang, R.; Zhang, S.; Zhang, S.; Zhang, W.; Zhang, W.; Zhang, X.; Zhang,
  X.; Zhao, H.; Zhao, Q.; Zhao, X.; Zhou, F.; Zhou, Z.; Zhuo, J.; Zou, Y.; Qiu,
  X.; Qiao, Y.; and Lin, D. 2024.
\newblock InternLM2 Technical Report.

\bibitem[{Chen et~al.(2023{\natexlab{a}})Chen, Zhu, Shen, Li, Liu, Zhang,
  Krishnamoorthi, Chandra, Xiong, and Elhoseiny}]{chen2023minigptv2}
Chen, J.; Zhu, D.; Shen, X.; Li, X.; Liu, Z.; Zhang, P.; Krishnamoorthi, R.;
  Chandra, V.; Xiong, Y.; and Elhoseiny, M. 2023{\natexlab{a}}.
\newblock MiniGPT-v2: large language model as a unified interface for
  vision-language multi-task learning.
\newblock \emph{arXiv preprint}.

\bibitem[{Chen et~al.(2024{\natexlab{a}})Chen, Thapa, Chalamala, Athiwaratkun,
  Song, and Zou}]{chen2024dragonfly}
Chen, K.; Thapa, R.; Chalamala, R.; Athiwaratkun, B.; Song, S.~L.; and Zou, J.
  2024{\natexlab{a}}.
\newblock Dragonfly: Multi-Resolution Zoom Supercharges Large Visual-Language
  Model.
\newblock \emph{arXiv preprint}.

\bibitem[{Chen et~al.(2024{\natexlab{b}})Chen, Li, Dong, Zhang, Zang, Chen,
  Duan, Wang, Qiao, Lin et~al.}]{chen2024we}
Chen, L.; Li, J.; Dong, X.; Zhang, P.; Zang, Y.; Chen, Z.; Duan, H.; Wang, J.;
  Qiao, Y.; Lin, D.; et~al. 2024{\natexlab{b}}.
\newblock Are We on the Right Way for Evaluating Large Vision-Language Models?
\newblock \emph{arXiv preprint}.

\bibitem[{Chen et~al.(2024{\natexlab{c}})Chen, Wang, Tian, Ye, Gao, Cui, Tong,
  Hu, Luo, Ma et~al.}]{chen2024far}
Chen, Z.; Wang, W.; Tian, H.; Ye, S.; Gao, Z.; Cui, E.; Tong, W.; Hu, K.; Luo,
  J.; Ma, Z.; et~al. 2024{\natexlab{c}}.
\newblock How far are we to gpt-4v? closing the gap to commercial multimodal
  models with open-source suites.
\newblock \emph{arXiv preprint}.

\bibitem[{Chen et~al.(2023{\natexlab{b}})Chen, Wu, Wang, Su, Chen, Xing, Zhong,
  Zhang, Zhu, Lu, Li, Luo, Lu, Qiao, and Dai}]{chen2023internvl}
Chen, Z.; Wu, J.; Wang, W.; Su, W.; Chen, G.; Xing, S.; Zhong, M.; Zhang, Q.;
  Zhu, X.; Lu, L.; Li, B.; Luo, P.; Lu, T.; Qiao, Y.; and Dai, J.
  2023{\natexlab{b}}.
\newblock InternVL: Scaling up Vision Foundation Models and Aligning for
  Generic Visual-Linguistic Tasks.
\newblock \emph{arXiv preprint}.

\bibitem[{Dong et~al.(2024{\natexlab{a}})Dong, Zhang, Zang, Cao, Wang, Ouyang,
  Wei, Zhang, Duan, Cao, Zhang, Li, Yan, Gao, Zhang, Li, Li, Chen, He, Zhang,
  Qiao, Lin, and Wang}]{internlmxcomposer2}
Dong, X.; Zhang, P.; Zang, Y.; Cao, Y.; Wang, B.; Ouyang, L.; Wei, X.; Zhang,
  S.; Duan, H.; Cao, M.; Zhang, W.; Li, Y.; Yan, H.; Gao, Y.; Zhang, X.; Li,
  W.; Li, J.; Chen, K.; He, C.; Zhang, X.; Qiao, Y.; Lin, D.; and Wang, J.
  2024{\natexlab{a}}.
\newblock InternLM-XComposer2: Mastering Free-form Text-Image Composition and
  Comprehension in Vision-Language Large Model.
\newblock \emph{arXiv preprint}.

\bibitem[{Dong et~al.(2024{\natexlab{b}})Dong, Zhang, Zang, Cao, Wang, Ouyang,
  Zhang, Duan, Zhang, Li, Yan, Gao, Chen, Zhang, Li, Li, Wang, Chen, He, Zhang,
  Dai, Qiao, Lin, and Wang}]{internlmxcomposer2_4khd}
Dong, X.; Zhang, P.; Zang, Y.; Cao, Y.; Wang, B.; Ouyang, L.; Zhang, S.; Duan,
  H.; Zhang, W.; Li, Y.; Yan, H.; Gao, Y.; Chen, Z.; Zhang, X.; Li, W.; Li, J.;
  Wang, W.; Chen, K.; He, C.; Zhang, X.; Dai, J.; Qiao, Y.; Lin, D.; and Wang,
  J. 2024{\natexlab{b}}.
\newblock InternLM-XComposer2-4KHD: A Pioneering Large Vision-Language Model
  Handling Resolutions from 336 Pixels to 4K HD.
\newblock \emph{arXiv preprint}.

\bibitem[{Dosovitskiy et~al.(2021)Dosovitskiy, Beyer, Kolesnikov, Weissenborn,
  Zhai, Unterthiner, Dehghani, Minderer, Heigold, Gelly, Uszkoreit, and
  Houlsby}]{dosovitskiy2020image}
Dosovitskiy, A.; Beyer, L.; Kolesnikov, A.; Weissenborn, D.; Zhai, X.;
  Unterthiner, T.; Dehghani, M.; Minderer, M.; Heigold, G.; Gelly, S.;
  Uszkoreit, J.; and Houlsby, N. 2021.
\newblock An Image is Worth 16x16 Words: Transformers for Image Recognition at
  Scale.
\newblock In \emph{ICLR}.

\bibitem[{Du et~al.(2022)Du, Qian, Liu, Ding, Qiu, Yang, and Tang}]{du2022glm}
Du, Z.; Qian, Y.; Liu, X.; Ding, M.; Qiu, J.; Yang, Z.; and Tang, J. 2022.
\newblock GLM: General Language Model Pretraining with Autoregressive Blank
  Infilling.
\newblock In \emph{ACL}.

\bibitem[{Duan et~al.(2024)Duan, Yang, Qiao, Fang, Chen, Liu, Dong, Zang,
  Zhang, Wang et~al.}]{duan2024vlmevalkit}
Duan, H.; Yang, J.; Qiao, Y.; Fang, X.; Chen, L.; Liu, Y.; Dong, X.; Zang, Y.;
  Zhang, P.; Wang, J.; et~al. 2024.
\newblock VLMEvalKit: An Open-Source Toolkit for Evaluating Large
  Multi-Modality Models.
\newblock \emph{arXiv preprint}.

\bibitem[{Fu et~al.(2023)Fu, Chen, Shen, Qin, Zhang, Lin, Yang, Zheng, Li, Sun
  et~al.}]{fu2023mme}
Fu, C.; Chen, P.; Shen, Y.; Qin, Y.; Zhang, M.; Lin, X.; Yang, J.; Zheng, X.;
  Li, K.; Sun, X.; et~al. 2023.
\newblock MME: A Comprehensive Evaluation Benchmark for Multimodal Large
  Language Models.
\newblock \emph{arXiv preprint}.

\bibitem[{Fu et~al.(2024)Fu, Hu, Li, Feng, Wang, Lin, Roth, Smith, Ma, and
  Krishna}]{fu2024blink}
Fu, X.; Hu, Y.; Li, B.; Feng, Y.; Wang, H.; Lin, X.; Roth, D.; Smith, N.~A.;
  Ma, W.-C.; and Krishna, R. 2024.
\newblock BLINK: Multimodal Large Language Models Can See but Not Perceive.
\newblock \emph{arXiv preprint}.

\bibitem[{Ge et~al.(2024)Ge, Cheng, Wang, Yuan, Gao, Song, Song, Huang, and
  Zheng}]{ge2024convllava}
Ge, C.; Cheng, S.; Wang, Z.; Yuan, J.; Gao, Y.; Song, J.; Song, S.; Huang, G.;
  and Zheng, B. 2024.
\newblock ConvLLaVA: Hierarchical Backbones as Visual Encoder for Large
  Multimodal Models.
\newblock \emph{arXiv preprint}.

\bibitem[{GLM et~al.(2024)GLM, Zeng, Xu, Wang, Zhang, Yin, Rojas, Feng, Zhao,
  Lai, Yu, Wang, Sun, Zhang, Cheng, Gui, Tang, Zhang, Li, Zhao, Wu, Zhong, Liu,
  Huang, Zhang, Zheng, Lu, Duan, Zhang, Cao, Yang, Tam, Zhao, Liu, Xia, Zhang,
  Gu, Lv, Liu, Liu, Yang, Song, Zhang, An, Xu, Niu, Yang, Li, Bai, Dong, Qi,
  Wang, Yang, Du, Hou, and Wang}]{glm2024chatglm}
GLM, T.; Zeng, A.; Xu, B.; Wang, B.; Zhang, C.; Yin, D.; Rojas, D.; Feng, G.;
  Zhao, H.; Lai, H.; Yu, H.; Wang, H.; Sun, J.; Zhang, J.; Cheng, J.; Gui, J.;
  Tang, J.; Zhang, J.; Li, J.; Zhao, L.; Wu, L.; Zhong, L.; Liu, M.; Huang, M.;
  Zhang, P.; Zheng, Q.; Lu, R.; Duan, S.; Zhang, S.; Cao, S.; Yang, S.; Tam,
  W.~L.; Zhao, W.; Liu, X.; Xia, X.; Zhang, X.; Gu, X.; Lv, X.; Liu, X.; Liu,
  X.; Yang, X.; Song, X.; Zhang, X.; An, Y.; Xu, Y.; Niu, Y.; Yang, Y.; Li, Y.;
  Bai, Y.; Dong, Y.; Qi, Z.; Wang, Z.; Yang, Z.; Du, Z.; Hou, Z.; and Wang, Z.
  2024.
\newblock ChatGLM: A Family of Large Language Models from GLM-130B to GLM-4 All
  Tools.

\bibitem[{Gu et~al.(2019)Gu, Lugmayr, Danelljan, Fritsche, Lamour, and
  Timofte}]{gu2019div8k}
Gu, S.; Lugmayr, A.; Danelljan, M.; Fritsche, M.; Lamour, J.; and Timofte, R.
  2019.
\newblock Div8k: Diverse 8k resolution image dataset.
\newblock In \emph{ICCVW}.

\bibitem[{Guan et~al.(2024)Guan, Liu, Wu, Xian, Li, Liu, Wang, Chen, Huang,
  Yacoob et~al.}]{guan2024hallusionbench}
Guan, T.; Liu, F.; Wu, X.; Xian, R.; Li, Z.; Liu, X.; Wang, X.; Chen, L.;
  Huang, F.; Yacoob, Y.; et~al. 2024.
\newblock HallusionBench: an advanced diagnostic suite for entangled language
  hallucination and visual illusion in large vision-language models.
\newblock In \emph{CVPR}.

\bibitem[{Han et~al.(2023)Han, You, Liu, Chen, Zheng, Mrini, Lin, Wang, Zhai,
  Yuan, Wang, and Yang}]{han2023coremm}
Han, X.; You, Q.; Liu, Y.; Chen, W.; Zheng, H.; Mrini, K.; Lin, X.; Wang, Y.;
  Zhai, B.; Yuan, J.; Wang, H.; and Yang, H. 2023.
\newblock CORE-MM: Complex Open-Ended Reasoning Evaluation For Multi-Modal
  Large Language Models.

\bibitem[{He et~al.(2016)He, Zhang, Ren, and Sun}]{he2016deep}
He, K.; Zhang, X.; Ren, S.; and Sun, J. 2016.
\newblock Deep residual learning for image recognition.
\newblock In \emph{CVPR}.

\bibitem[{Izacard et~al.(2022)Izacard, Caron, Hosseini, Riedel, Bojanowski,
  Joulin, and Grave}]{izacard2022unsupervised}
Izacard, G.; Caron, M.; Hosseini, L.; Riedel, S.; Bojanowski, P.; Joulin, A.;
  and Grave, E. 2022.
\newblock Unsupervised Dense Information Retrieval with Contrastive Learning.
\newblock \emph{TMLR}.

\bibitem[{Kembhavi et~al.(2016)Kembhavi, Salvato, Kolve, Seo, Hajishirzi, and
  Farhadi}]{Kembhavi2016ADI}
Kembhavi, A.; Salvato, M.; Kolve, E.; Seo, M.; Hajishirzi, H.; and Farhadi, A.
  2016.
\newblock A Diagram is Worth a Dozen Images.
\newblock In \emph{ECCV}.

\bibitem[{Kirillov et~al.(2023)Kirillov, Mintun, Ravi, Mao, Rolland, Gustafson,
  Xiao, Whitehead, Berg, Lo et~al.}]{kirillov2023segment}
Kirillov, A.; Mintun, E.; Ravi, N.; Mao, H.; Rolland, C.; Gustafson, L.; Xiao,
  T.; Whitehead, S.; Berg, A.~C.; Lo, W.-Y.; et~al. 2023.
\newblock Segment anything.
\newblock In \emph{ICCV}.

\bibitem[{Lauren\c{c}on et~al.(2023)Lauren\c{c}on, Saulnier, Tronchon, Bekman,
  Singh, Lozhkov, Wang, Karamcheti, Rush, Kiela, Cord, and
  Sanh}]{laurencon2023obelics}
Lauren\c{c}on, H.; Saulnier, L.; Tronchon, L.; Bekman, S.; Singh, A.; Lozhkov,
  A.; Wang, T.; Karamcheti, S.; Rush, A.; Kiela, D.; Cord, M.; and Sanh, V.
  2023.
\newblock OBELICS: An Open Web-Scale Filtered Dataset of Interleaved Image-Text
  Documents.

\bibitem[{LeCun et~al.(1989)LeCun, Boser, Denker, Henderson, Howard, Hubbard,
  and Jackel}]{lecun1989backpropagation}
LeCun, Y.; Boser, B.; Denker, J.~S.; Henderson, D.; Howard, R.~E.; Hubbard, W.;
  and Jackel, L.~D. 1989.
\newblock Backpropagation applied to handwritten zip code recognition.
\newblock \emph{Neural computation}, 1.

\bibitem[{Li et~al.(2024{\natexlab{a}})Li, Ge, Ge, Wang, Wang, Zhang, and
  Shan}]{li2023seed2}
Li, B.; Ge, Y.; Ge, Y.; Wang, G.; Wang, R.; Zhang, R.; and Shan, Y.
  2024{\natexlab{a}}.
\newblock SEED-Bench: Benchmarking Multimodal Large Language Models.
\newblock In \emph{CVPR}, 13299--13308.

\bibitem[{Li et~al.(2024{\natexlab{b}})Li, Zhang, Wang, Zhong, Chen, Chu, Liu,
  and Jia}]{li2024mini}
Li, Y.; Zhang, Y.; Wang, C.; Zhong, Z.; Chen, Y.; Chu, R.; Liu, S.; and Jia, J.
  2024{\natexlab{b}}.
\newblock Mini-gemini: Mining the potential of multi-modality vision language
  models.
\newblock \emph{arXiv preprint}.

\bibitem[{Li et~al.(2024{\natexlab{c}})Li, Yang, Liu, Ma, Zhang, Yang, Sun,
  Liu, and Bai}]{li2024monkey}
Li, Z.; Yang, B.; Liu, Q.; Ma, Z.; Zhang, S.; Yang, J.; Sun, Y.; Liu, Y.; and
  Bai, X. 2024{\natexlab{c}}.
\newblock Monkey: Image resolution and text label are important things for
  large multi-modal models.
\newblock In \emph{CVPR}.

\bibitem[{Liu et~al.(2024{\natexlab{a}})Liu, Li, Li, and Lee}]{liu2024improved}
Liu, H.; Li, C.; Li, Y.; and Lee, Y.~J. 2024{\natexlab{a}}.
\newblock Improved baselines with visual instruction tuning.
\newblock In \emph{CVPR}.

\bibitem[{Liu et~al.(2024{\natexlab{b}})Liu, Li, Li, Li, Zhang, Shen, and
  Lee}]{liu2024llavanext}
Liu, H.; Li, C.; Li, Y.; Li, B.; Zhang, Y.; Shen, S.; and Lee, Y.~J.
  2024{\natexlab{b}}.
\newblock LLaVA-NeXT: Improved reasoning, OCR, and world knowledge.

\bibitem[{Liu et~al.(2024{\natexlab{c}})Liu, You, Han, Wang, Zhai, Liu, Tao,
  Huang, He, and Yang}]{liu2024infimm}
Liu, H.; You, Q.; Han, X.; Wang, Y.; Zhai, B.; Liu, Y.; Tao, Y.; Huang, H.; He,
  R.; and Yang, H. 2024{\natexlab{c}}.
\newblock InfiMM-HD: A Leap Forward in High-Resolution Multimodal
  Understanding.
\newblock \emph{arXiv preprint}.

\bibitem[{Liu et~al.(2024{\natexlab{d}})Liu, Li, Yang, Li, Yin, lin Liu, Jin,
  and Bai}]{liu2024hidden}
Liu, Y.; Li, Z.; Yang, B.; Li, C.; Yin, X.; lin Liu, C.; Jin, L.; and Bai, X.
  2024{\natexlab{d}}.
\newblock On the Hidden Mystery of OCR in Large Multimodal Models.

\bibitem[{Liu et~al.(2022)Liu, Mao, Wu, Feichtenhofer, Darrell, and
  Xie}]{liu2022convnet}
Liu, Z.; Mao, H.; Wu, C.-Y.; Feichtenhofer, C.; Darrell, T.; and Xie, S. 2022.
\newblock A convnet for the 2020s.
\newblock In \emph{CVPR}.

\bibitem[{Lu et~al.(2024{\natexlab{a}})Lu, Liu, Zhang, Wang, Dong, Liu, Sun,
  Ren, Li, Yang, Sun, Deng, Xu, Xie, and Ruan}]{lu2024deepseekvl}
Lu, H.; Liu, W.; Zhang, B.; Wang, B.; Dong, K.; Liu, B.; Sun, J.; Ren, T.; Li,
  Z.; Yang, H.; Sun, Y.; Deng, C.; Xu, H.; Xie, Z.; and Ruan, C.
  2024{\natexlab{a}}.
\newblock DeepSeek-VL: Towards Real-World Vision-Language Understanding.

\bibitem[{Lu et~al.(2024{\natexlab{b}})Lu, Bansal, Xia, Liu, Li, Hajishirzi,
  Cheng, Chang, Galley, and Gao}]{lu2024mathvista}
Lu, P.; Bansal, H.; Xia, T.; Liu, J.; Li, C.; Hajishirzi, H.; Cheng, H.; Chang,
  K.-W.; Galley, M.; and Gao, J. 2024{\natexlab{b}}.
\newblock MathVista: Evaluating Mathematical Reasoning of Foundation Models in
  Visual Contexts.
\newblock In \emph{ICLR}.

\bibitem[{Lu et~al.(2022)Lu, Mishra, Xia, Qiu, Chang, Zhu, Tafjord, Clark, and
  Kalyan}]{lu2022learn}
Lu, P.; Mishra, S.; Xia, T.; Qiu, L.; Chang, K.-W.; Zhu, S.-C.; Tafjord, O.;
  Clark, P.; and Kalyan, A. 2022.
\newblock Learn to Explain: Multimodal Reasoning via Thought Chains for Science
  Question Answering.
\newblock In \emph{NeurIPS}.

\bibitem[{Luo et~al.(2024)Luo, Zhou, Zhang, Zheng, Sun, and Ji}]{luo2024feast}
Luo, G.; Zhou, Y.; Zhang, Y.; Zheng, X.; Sun, X.; and Ji, R. 2024.
\newblock Feast Your Eyes: Mixture-of-Resolution Adaptation for Multimodal
  Large Language Models.
\newblock \emph{arXiv preprint}.

\bibitem[{Ma et~al.(2024)Ma, Liu, Wong, Rao, Li, Ding, Chao, Tao, and
  Zhang}]{ma-etal-2024-3am}
Ma, X.; Liu, X.; Wong, D.~F.; Rao, J.; Li, B.; Ding, L.; Chao, L.~S.; Tao, D.;
  and Zhang, M. 2024.
\newblock 3{AM}: An Ambiguity-Aware Multi-Modal Machine Translation Dataset.
\newblock In Calzolari, N.; Kan, M.-Y.; Hoste, V.; Lenci, A.; Sakti, S.; and
  Xue, N., eds., \emph{COLING}.

\bibitem[{Masry et~al.(2022)Masry, Long, Tan, Joty, and
  Hoque}]{masry-etal-2022-chartqa}
Masry, A.; Long, D.; Tan, J.~Q.; Joty, S.; and Hoque, E. 2022.
\newblock {C}hart{QA}: A Benchmark for Question Answering about Charts with
  Visual and Logical Reasoning.
\newblock In \emph{ACL}.

\bibitem[{Mathew, Karatzas, and Jawahar(2021)}]{mathew2021docvqa}
Mathew, M.; Karatzas, D.; and Jawahar, C. 2021.
\newblock Docvqa: A dataset for vqa on document images.
\newblock In \emph{WACV}.

\bibitem[{Mishra et~al.(2019)Mishra, Shekhar, Singh, and
  Chakraborty}]{mishraICDAR19}
Mishra, A.; Shekhar, S.; Singh, A.~K.; and Chakraborty, A. 2019.
\newblock OCR-VQA: Visual Question Answering by Reading Text in Images.
\newblock In \emph{ICDAR}.

\bibitem[{Radford et~al.(2021)Radford, Kim, Hallacy, Ramesh, Goh, Agarwal,
  Sastry, Askell, Mishkin, Clark et~al.}]{radford2021learning}
Radford, A.; Kim, J.~W.; Hallacy, C.; Ramesh, A.; Goh, G.; Agarwal, S.; Sastry,
  G.; Askell, A.; Mishkin, P.; Clark, J.; et~al. 2021.
\newblock Learning transferable visual models from natural language
  supervision.
\newblock In \emph{ICML}.

\bibitem[{Reid et~al.(2024)Reid, Savinov, Teplyashin, Lepikhin, Lillicrap,
  Alayrac, Soricut, Lazaridou, Firat, Schrittwieser et~al.}]{reid2024gemini}
Reid, M.; Savinov, N.; Teplyashin, D.; Lepikhin, D.; Lillicrap, T.; Alayrac,
  J.-b.; Soricut, R.; Lazaridou, A.; Firat, O.; Schrittwieser, J.; et~al. 2024.
\newblock Gemini 1.5: Unlocking multimodal understanding across millions of
  tokens of context.
\newblock \emph{arXiv preprint}.

\bibitem[{Singh et~al.(2019)Singh, Natarjan, Shah, Jiang, Chen, Parikh, and
  Rohrbach}]{singh2019towards}
Singh, A.; Natarjan, V.; Shah, M.; Jiang, Y.; Chen, X.; Parikh, D.; and
  Rohrbach, M. 2019.
\newblock Towards VQA Models That Can Read.
\newblock In \emph{CVPR}.

\bibitem[{Team(2024)}]{team2024chameleon}
Team, C. 2024.
\newblock Chameleon: Mixed-modal early-fusion foundation models.
\newblock \emph{arXiv preprint}.

\bibitem[{Tong et~al.(2024)Tong, Liu, Zhai, Ma, LeCun, and Xie}]{tong2024eyes}
Tong, S.; Liu, Z.; Zhai, Y.; Ma, Y.; LeCun, Y.; and Xie, S. 2024.
\newblock Eyes wide shut? exploring the visual shortcomings of multimodal llms.
\newblock In \emph{CVPR}.

\bibitem[{Touvron et~al.(2023{\natexlab{a}})Touvron, Lavril, Izacard, Martinet,
  Lachaux, Lacroix, Rozi{\`e}re, Goyal, Hambro, Azhar
  et~al.}]{touvron2023llama}
Touvron, H.; Lavril, T.; Izacard, G.; Martinet, X.; Lachaux, M.-A.; Lacroix,
  T.; Rozi{\`e}re, B.; Goyal, N.; Hambro, E.; Azhar, F.; et~al.
  2023{\natexlab{a}}.
\newblock Llama: Open and efficient foundation language models.
\newblock \emph{arXiv preprint}.

\bibitem[{Touvron et~al.(2023{\natexlab{b}})Touvron, Martin, Stone, Albert,
  Almahairi, Babaei, Bashlykov, Batra, Bhargava, Bhosale
  et~al.}]{touvron2023llama2}
Touvron, H.; Martin, L.; Stone, K.; Albert, P.; Almahairi, A.; Babaei, Y.;
  Bashlykov, N.; Batra, S.; Bhargava, P.; Bhosale, S.; et~al.
  2023{\natexlab{b}}.
\newblock Llama 2: Open foundation and fine-tuned chat models.
\newblock \emph{arXiv preprint}.

\bibitem[{Wang et~al.(2023)Wang, Lv, Yu, Hong, Qi, Wang, Ji, Yang, Zhao, Song,
  Xu, Xu, Li, Dong, Ding, and Tang}]{wang2023cogvlm}
Wang, W.; Lv, Q.; Yu, W.; Hong, W.; Qi, J.; Wang, Y.; Ji, J.; Yang, Z.; Zhao,
  L.; Song, X.; Xu, J.; Xu, B.; Li, J.; Dong, Y.; Ding, M.; and Tang, J. 2023.
\newblock CogVLM: Visual Expert for Pretrained Language Models.

\bibitem[{Wei et~al.(2023)Wei, Kong, Chen, Zhao, Ge, Yang, Sun, Han, and
  Zhang}]{wei2023vary}
Wei, H.; Kong, L.; Chen, J.; Zhao, L.; Ge, Z.; Yang, J.; Sun, J.; Han, C.; and
  Zhang, X. 2023.
\newblock Vary: Scaling up the vision vocabulary for large vision-language
  models.
\newblock \emph{arXiv preprint}.

\bibitem[{Wu and Xie(2024)}]{wu2024v}
Wu, P.; and Xie, S. 2024.
\newblock V*: Guided Visual Search as a Core Mechanism in Multimodal LLMs.
\newblock In \emph{CVPR}.

\bibitem[{Xu et~al.(2023)Xu, Ye, Yan, Shi, Ye, Xu, Li, Bi, Qian, Wang, Xu,
  Zhang, Huang, Huang, and Zhou}]{Xu2023mPLUG2AM}
Xu, H.; Ye, Q.; Yan, M.; Shi, Y.; Ye, J.; Xu, Y.; Li, C.; Bi, B.; Qian, Q.;
  Wang, W.; Xu, G.; Zhang, J.; Huang, S.; Huang, F.; and Zhou, J. 2023.
\newblock m{PLUG}-2: A Modularized Multi-modal Foundation Model Across Text,
  Image and Video.
\newblock In \emph{ICML}.

\bibitem[{Yifan et~al.(2023)Yifan, Yifan, Kun, Jinpeng, Xin, and
  Ji-Rong}]{li2023evaluating}
Yifan, L.; Yifan, D.; Kun, Z.; Jinpeng, W.; Xin, Z.; and Ji-Rong, W. 2023.
\newblock Evaluating Object Hallucination in Large Vision-Language Models.
\newblock In \emph{EMNLP}.

\bibitem[{Young et~al.(2024)Young, Chen, Li, Huang, Zhang, Zhang, Li, Zhu,
  Chen, Chang et~al.}]{young2024yi}
Young, A.; Chen, B.; Li, C.; Huang, C.; Zhang, G.; Zhang, G.; Li, H.; Zhu, J.;
  Chen, J.; Chang, J.; et~al. 2024.
\newblock Yi: Open foundation models by 01. ai.
\newblock \emph{arXiv preprint}.

\bibitem[{Yu et~al.(2024)Yu, Yang, Li, Wang, Lin, Liu, Wang, and
  Wang}]{yu2024mm}
Yu, W.; Yang, Z.; Li, L.; Wang, J.; Lin, K.; Liu, Z.; Wang, X.; and Wang, L.
  2024.
\newblock Mm-vet: Evaluating large multimodal models for integrated
  capabilities.
\newblock In \emph{ICML}.

\bibitem[{Yuan et~al.(2023)Yuan, Haodong, Yuanhan, Bo, Songyang, Wangbo, Yike,
  Jiaqi, Conghui, Liu, Kai, and Dahua}]{MMBench}
Yuan, L.; Haodong, D.; Yuanhan, Z.; Bo, L.; Songyang, Z.; Wangbo, Z.; Yike, Y.;
  Jiaqi, W.; Conghui, H.; Liu, Z.; Kai, C.; and Dahua, L. 2023.
\newblock MMBench: Is Your Multi-modal Model an All-around Player?
\newblock \emph{arXiv preprint}.

\bibitem[{Yue et~al.(2024)Yue, Ni, Zhang, Zheng, Liu, Zhang, Stevens, Jiang,
  Ren, Sun, Wei, Yu, Yuan, Sun, Yin, Zheng, Yang, Liu, Huang, Sun, Su, and
  Chen}]{yue2023mmmu}
Yue, X.; Ni, Y.; Zhang, K.; Zheng, T.; Liu, R.; Zhang, G.; Stevens, S.; Jiang,
  D.; Ren, W.; Sun, Y.; Wei, C.; Yu, B.; Yuan, R.; Sun, R.; Yin, M.; Zheng, B.;
  Yang, Z.; Liu, Y.; Huang, W.; Sun, H.; Su, Y.; and Chen, W. 2024.
\newblock MMMU: A Massive Multi-discipline Multimodal Understanding and
  Reasoning Benchmark for Expert AGI.
\newblock In \emph{CVPR}.

\bibitem[{Zhang et~al.(2023)Zhang, Dong, Wang, Cao, Xu, Ouyang, Zhao, Ding,
  Zhang, Duan, Zhang, Yan, Zhang, Li, Li, Chen, He, Zhang, Qiao, Lin, and
  Wang}]{zhang2023internlmxcomposer}
Zhang, P.; Dong, X.; Wang, B.; Cao, Y.; Xu, C.; Ouyang, L.; Zhao, Z.; Ding, S.;
  Zhang, S.; Duan, H.; Zhang, W.; Yan, H.; Zhang, X.; Li, W.; Li, J.; Chen, K.;
  He, C.; Zhang, X.; Qiao, Y.; Lin, D.; and Wang, J. 2023.
\newblock InternLM-XComposer: A Vision-Language Large Model for Advanced
  Text-image Comprehension and Composition.

\bibitem[{Zhang et~al.(2024)Zhang, Wen, Fu, Wang, Zhang, Wang, and
  Jin}]{zhang2024beyond}
Zhang, Y.-F.; Wen, Q.; Fu, C.; Wang, X.; Zhang, Z.; Wang, L.; and Jin, R. 2024.
\newblock Beyond LLaVA-HD: Diving into High-Resolution Large Multimodal Models.
\newblock \emph{arXiv preprint}.

\bibitem[{Zheng et~al.(2023)Zheng, Zhou, Meng, Zhou, and
  Huang}]{zheng2023large}
Zheng, C.; Zhou, H.; Meng, F.; Zhou, J.; and Huang, M. 2023.
\newblock Large language models are not robust multiple choice selectors.
\newblock In \emph{ICLR}.

\bibitem[{Zhou et~al.(2024)Zhou, Cui, Yoon, Zhang, Deng, Finn, Bansal, and
  Yao}]{zhouanalyzing}
Zhou, Y.; Cui, C.; Yoon, J.; Zhang, L.; Deng, Z.; Finn, C.; Bansal, M.; and
  Yao, H. 2024.
\newblock Analyzing and Mitigating Object Hallucination in Large
  Vision-Language Models.
\newblock In \emph{ICLR}.

\end{thebibliography}

\clearpage
\appendix
\section{A. HR-Bench}
\label{sec:hrbench}
\subsection{Curation Procedure}
Our data curation process is conducted in a semi-automatic manner. The procedure consists of three main steps:

We observe that even the current SOTA MLLMs~\cite{achiam2023gpt,reid2024gemini} are unable to accurately perceive objects that appear in 8K images. Therefore, we first manually annotate the bounding boxes around the objects in the images.

Second, we crop the images within the bounding boxes and use GPT-4o to generate the query and answer pairs for the objects in the images.

Third, to ensure the highest quality benchmark, we involve human experts to meticulously review and filter out any incorrect or ambiguous queries after generating the query and answer pairs.

Following this three-stage process, we finalize the benchmark, which results in a total of 200 image queries with improved accuracy and reliability. Subsequently, we will further analyze our \itbf{HR-Bench}, including \textbf{benchmark statistics}, \textbf{evaluation metrics}, \textbf{multi-modal gain} and \textbf{data leakage}.

\subsection{Benchmark Statistics}
As detailed in Table~\ref{table:hr_bench_info}, \itbf{HR-Bench} consists two sub-tasks \itbf{Fine-grained Single-instance Perception (FSP)}, which includes tasks such as attribute recognition, OCR, visual prompting, and \itbf{Fine-grained Cross-instance Perception (FCP)}, which encompasses map analysis, chart analysis, and spatial relationship assessment.

\begin{table*}[thbp]
\begin{tabular}{p{0.1\linewidth}p{0.2\linewidth}p{0.5\linewidth}p{0.1\linewidth}}
\toprule
\textbf{Type}                 & \textbf{Task}                  & \textbf{Description} & \textbf{\# Samples} \\ \hline
\multirow{3}{*}{\itbf{FSP}} & Attribute Recognition  &     Determine the instance's attributes, such as color, shape, or material        &       \multirow{3}{*}{100}     \\ \cmidrule(lr){2-3}
                     & OCR                   &         MLLM should answer questions about textual elements in the image.    &            \\ \cmidrule(lr){2-3}
                     & Visual Prompting         &         The visual prompting allows highlighting specific areas within images, facilitating the assessment of MLLMs' detailed comprehension of these regions    &            \\ \hline
\multirow{3}{*}{\itbf{FCP}} & Map Analysis          &     MLLM need to combine relevant geographical knowledge to infer whether the two points in the image belong to the same country.        &    \multirow{3}{*}{100}        \\ \cmidrule(lr){2-3}
                     & Chart Analysis                &         Evaluating MLLM ability to understand and extract information from chart.    &            \\ \cmidrule(lr){2-3}
                     & Spatial Relationship  &         MLLM should ground two mentioned objects and recognize their relative spatial relation within the image.    &      \\
\bottomrule
\end{tabular}
\caption{Breakdown of the \itbf{FSP} and \itbf{FCP} tasks evaluated in the \itbf{HR-Bench}. The examples are sourced from DIV8K and Internet.}
\label{table:hr_bench_info}
\end{table*}

We further compare 21 widely used MLLM benchmarks with our \itbf{HR-Bench}. As shown in Table~\ref{table:compare_benchmark}, we find that 1) the highest resolution among current MLLM benchmarks is only 2K, with a lack of 4K and higher resolution benchmark; and 2) there are currently few MLLM benchmarks that involve visual prompting and map analysis tasks. 
\begin{table*}[hbt]
\centering
\setlength{\tabcolsep}{4.2mm}
\begin{tabular}{lccccccc}
\toprule
Benchmark & Average Resolution & AR & OCR & VP & MA & CA & SR \\ \hline
      MMVP~\cite{tong2024eyes}    &  224   &  \green{\checkmark}  &  \green{\checkmark}  & \red{\ding{55}}  & \red{\ding{55}}   & \red{\ding{55}}  & \green{\checkmark}  \\
      MMBench~\cite{MMBench}    &   450  &  \green{\checkmark}  &  \green{\checkmark} &  \red{\ding{55}}  &  \green{\checkmark}  & \green{\checkmark}  &  \green{\checkmark}       \\
      CCBench    &  450   &  \green{\checkmark}  &  \red{\ding{55}}  &  \red{\ding{55}}  &  \red{\ding{55}}  & \red{\ding{55}}  &   \red{\ding{55}}       \\
      OCRVQA~\cite{mishraICDAR19}    &  493   &  \green{\checkmark}  &  \green{\checkmark}  &  \red{\ding{55}}  &  \red{\ding{55}}  & \red{\ding{55}}  &    \red{\ding{55}}      \\
      MMVet~\cite{yu2024mm}    &  494   &  \green{\checkmark}  & \green{\checkmark}   &  \red{\ding{55}}  &  \green{\checkmark}  & \green{\checkmark}  &   \green{\checkmark}      \\
      MME~\cite{fu2023mme}    &  494   &  \green{\checkmark}  &  \green{\checkmark}  & \green{\checkmark}   &  \red{\ding{55}}  & \green{\checkmark}  &  \green{\checkmark}        \\
      ScienceQA~\cite{lu2022learn}    &  501   &  \green{\checkmark}  &  \green{\checkmark}  &   \red{\ding{55}} &  \green{\checkmark}  & \green{\checkmark}  &      \green{\checkmark}   \\
      SEEDBench~\cite{li2023seed2}    &  512   &  \green{\checkmark}  &  \green{\checkmark}  &  \green{\checkmark}  &  \green{\checkmark}  & \green{\checkmark}  & \green{\checkmark}         \\
      MMStar~\cite{chen2024we}    &  534   & \green{\checkmark}   &  \green{\checkmark}  &  \red{\ding{55}}  &  \green{\checkmark}  &  \green{\checkmark} &   \green{\checkmark}       \\
      MathVista~\cite{lu2024mathvista}    &  603   &  \green{\checkmark}  &  \green{\checkmark}  &  \red{\ding{55}}  &  \green{\checkmark}  & \green{\checkmark}  &     \green{\checkmark}    \\
      POPE~\cite{li2023evaluating}    &   617  &  \green{\checkmark}  &  \red{\ding{55}}  &  \red{\ding{55}}  & \red{\ding{55}}   &  \red{\ding{55}} &   \red{\ding{55}}       \\
      AI2D~\cite{Kembhavi2016ADI}    &  653   &  \green{\checkmark}  &  \green{\checkmark}  &  \red{\ding{55}}  &  \green{\checkmark}  &  \green{\checkmark} &    \green{\checkmark}     \\
      MMMU~\cite{yue2023mmmu}    &  691   &  \green{\checkmark}  &  \green{\checkmark}  &  \red{\ding{55}}  &  \green{\checkmark}  & \green{\checkmark}  &    \green{\checkmark}   \\
      ChartQA~\cite{masry-etal-2022-chartqa}    &   763  &  \red{\ding{55}}  &  \red{\ding{55}}  & \red{\ding{55}}   &  \red{\ding{55}}  & \green{\checkmark}  &   \red{\ding{55}}      \\
      OCRBench~\cite{liu2024hidden}    &  861   &  \green{\checkmark}  & \green{\checkmark}   &  \red{\ding{55}}  & \red{\ding{55}}   & \green{\checkmark}  & \green{\checkmark}  \\
      Blink~\cite{fu2024blink}    &  913   &  \green{\checkmark}  &  \red{\ding{55}}  &  \green{\checkmark}  &  \red{\ding{55}}  & \red{\ding{55}}  &   \green{\checkmark}    \\
      TextVQA~\cite{singh2019towards}    &  1023   & \green{\checkmark}   &  \green{\checkmark}  &  \red{\ding{55}}  & \red{\ding{55}}   & \red{\ding{55}}  &    \green{\checkmark}     \\
      HallusionBench~\cite{guan2024hallusionbench}    &  1202   &  \green{\checkmark}  &  \green{\checkmark}  &  \red{\ding{55}}  &  \green{\checkmark}  & \green{\checkmark}  &   \green{\checkmark}   \\
      RealWorldQA    &   1410  &  \green{\checkmark}  &  \green{\checkmark}  &  \red{\ding{55}}  &  \red{\ding{55}}  & \red{\ding{55}}  &  \green{\checkmark}    \\
      LLaVABench    &  1465   & \green{\checkmark}   & \green{\checkmark}   &  \red{\ding{55}}  &  \red{\ding{55}}  &  \red{\ding{55}} &    \green{\checkmark}  \\
      V$^*$~\cite{wu2024v}      &  2348   &  \green{\checkmark}  &  \green{\checkmark}  &  \red{\ding{55}}  &  \red{\ding{55}}  &  \red{\ding{55}} &  \green{\checkmark}     \\
      \hdashline
      \itbf{HR-Bench 4K}   &   \textbf{4032}  &  \green{\checkmark}  &  \green{\checkmark}  & \green{\checkmark}   &  \green{\checkmark}  & \green{\checkmark}  &  \green{\checkmark}        \\
      \itbf{HR-Bench 8K}   &   \textbf{7680}  &  \green{\checkmark}  &  \green{\checkmark}  &  \green{\checkmark}  &  \green{\checkmark}  & \green{\checkmark}  &  \green{\checkmark}    \\
\bottomrule
\end{tabular}
\caption{Comparisons between existing 22 MLLM benchmarks. We compare across seven dimensions: average resolution, AR (attrubute recognition), OCR, VP (visual prompting), MA (map analysis), CA (chart analysis), SR (spatial relationship). The current MLLM benchmarks have a maximum resolution of only 2K. Our proposed HR-Bench addresses the lack of HR images in this field. }
\label{table:compare_benchmark}
\end{table*}
\subsection{Benchmark Evaluation}
To address the sensitivity of MLLMs to the order of options in multiple-choice questions, we adopt a more robust evaluation method called \textbf{Cyclic Permutation}~\cite{zheng2023large}. Specifically, for a set of $n$ options, we reorder them by shifting their positions in a circular fashion. In this process, each item in the original sequence is moved to the position of the following item, with the last item looping back to the first position. As a result, $n$ computations are required for each sample. Finally, we calculate the average accuracy (\textit{ACC}) of the $n$ results:
\begin{equation}
    ACC = \frac{ACC_1+ACC_2+...+ACC_n}{n}.
\end{equation}

\subsection{Multi-modal Gain \& Multi-modal Leakage}
To ensure that our \itbf{HR-Bench} accurately reflects the actual performance gains of MLLMs derived from multi-modal training and mitigates the risk of data leakage, we draw inspiration from~\citet{chen2024we}. We compute the multi-modal gain (MG) and multi-modal leakage (ML) to assess the actual performance improvements from the multi-modal training process and quantify the extent of data leakage. 

To calculate the MG metric for a given MLLM on our \itbf{HR-Bench}, we measure the model's accuracy with visual inputs ($S_v$) and without visual inputs ($S_{wv}$). Then the MG metric is then derived using the following formula:
\begin{equation}
    MG=S_v-S_{wv}.
\end{equation}

To compute the ML metric for a given MLLM on our \itbf{HR-Bench}, we evaluate the given MLLM's LLM base (without any multi-modal training), denoted as $S_t$. Then the ML metric is formulated as follows:
\begin{equation}
    ML = max(0,S_{wv}-S_t).
\end{equation}

As seen in Table~\ref{table:mg_ml}, we illustrate the MG/ML metrics for each MLLM across 2 benchmarks. In the final row of table, we list the average multimodal gain and multi-modal leakage for existing MLLMs across 2 benchmarks for analysis. The MG on our \itbf{HR-Bench 8K} is relatively low due to the current MLLMs (\ie $S_v$) exhibiting low accuracy on this benchmark, leading to a reduced MG value. Additionally, \itbf{HR-Bench 8K} has the low degree of multi-modal leakage at 3.6. This provides a comprehensive arena for comparing existing MLLMs.

\begin{table}[htb]
\centering
\begin{tabular}{lcccc}
\toprule
\multirow{2}{*}{\textbf{Model}} & \multicolumn{2}{c}{\itbf{MMBench}}  & \multicolumn{2}{c}{\itbf{HR-Bench 8K}} \\  \cmidrule(lr){2-3} \cmidrule(lr){4-5}
                       & MG$\uparrow$          & ML$\downarrow$      & MG$\uparrow$          & ML$\downarrow$         \\ \hline
InternVL-1.5-26B       & \textbf{57.1}        & \textbf{9.7}           & \textbf{22.4}        & \textbf{0.0}        \\
LLaVA-1.6-34B          & 54.7         & 13.9            & 12.6        & 3.5        \\
LLaVA-HR-X-13B         & 43.2        & 15.6             & 17.3        & 3.6        \\
LLaVA-HR-X-7B          & 46.1         & 13.9            & 10.1        & 3.5        \\
LLaVA-1.6-7B          & 47.3        & 13.7              & 5.9         & 6.9        \\
Yi-VL-34B              & 49.1        & 11.3            & 3.4         & 4.3        \\
Yi-VL-6B               & 45.4         & 13.7            & 5.3         & \textbf{0.0}        \\
LLaVA-v1.5-13B         & 47.0         & 14.3            & 5.0         & 6.8        \\
\hdashline
Avg. across models  &  48.7 & 13.3  & 10.3  & 3.6  \\
\bottomrule
\end{tabular}
\caption{Evaluation of various MLLMs on 2 Benchmarks with multi-modal gain (MG) and multi-modal leakage (ML) metrics. We present the results for 8 open-source MLLMs of varying sizes and architectures. The bottom row displays the average performance across all models. The \textbf{best} results are emphasized in \textbf{bold}.}
\label{table:mg_ml}
\end{table}

\section{B. Implementation Details of \textbf{DC$^2$}}
\label{sec:implement_method}
\subsection{Prompt Templates for Conquer Stage}
\label{sec:conquer}
In the conquer stage, we utilize MLLM with different prompts (\ie $\mathcal{F}_{leaf}$ and $\mathcal{F}_{non-leaf}$) to generate the text descriptions and extract objects. The prompts used in the conquer stage are listed in Table~\ref{table:conquer_prompt}.
\begin{table*}[thb]
\begin{tabularx}{\linewidth}{cX}
\toprule
\multicolumn{2}{l}{\textbf{Prompt Templates for Conquer Stage}}     \\ \hline
\multirow{1}{*}[-0.5ex]{Prompt for Leaf Node} & Please describe this image. \\
\midrule
\multirow{3}{*}[-0.0ex]{Prompt for Non-leaf Node} & Give you patch captions which describe the image patches repectively, you are required to combine all the information to generate refined text descriptions about the image. Patch Captions: 
\{Text Descriptions of Image Patches\}\\ \hline
\multirow{3}{*}[-12.0ex]{Prompt for Key Objects Extraction} & \# System message\\
& You are a language assistant that helps to extract information from given sentences. Given a sentence which describe the image, extract the existent entities within the sentence for me. Extract the common objects and summarize them as general categories without repetition, merge essentially similar objects. Avoid extracting abstract or non-specific entities. Only extract concrete, certainly existent objects that fall in general categories and are described in a certain tone in the sentence. Extract entity in a JSON DICT. Output all the extracted types of items in one line and separate each object type with a period. You should ignore the singular and plural forms of nouns, and all extracted objects should be represented in singular form. If there is noting to output, then output a single empty list [].\\
& \# Examples \\
& Input: \\
& Sentence: ``The bus is surrounded by a few other vehicles, including a car and a truck, which are driving in the same direction as the bus. A person can be seen standing on the sidewalk, possibly waiting for the bus or observing the scene.''\\
& Output: \{``object\_list'': [``bus'',``car'',``truck'',``person'']\} \\
& Input: \\
& Sentence: ``\{Text Description\}''\\
& Output: \\
\bottomrule
\end{tabularx}
\caption{Prompt Templates for Conquer.}
\label{table:conquer_prompt}
\end{table*}

\subsection{Prompt Templates for Inference}
In the inference stage, we use the user prompt to interact with visual memory $\mathcal{M}$. Specifically, given a user prompt, we use Contriever-MSMARCO~\cite{izacard2022unsupervised} to retrieve related objects with confidence levels exceed $\alpha$. Subsequently, we obtain the image patches for the retrieved objects. Then, we utilize MLLM with prompt (as shown in Table~\ref{table:inference_prompt}) to generate accurate text descriptions. Finally, we concatenate the text descriptions with user prompt and utilize the MLLM to generate the final response.

\begin{table}[hbt]
\begin{tabularx}{\linewidth}{X}
\toprule
\textbf{Prompt Template for Inference}     \\ \hline
Question: \textbf{[question]} \\
You should provide more information to help you answer the question and explain the reasons. If no any helpful information, you should answer NONE.\\
\bottomrule
\end{tabularx}%
\caption{Prompt template for inference. The ``\textbf{[question]}'' is placeholder meant to be replaced with specific question from the dataset. For multiple-choice questions, we do not include the options in the \textbf{[question]}.}
\label{table:inference_prompt}
\end{table}

\subsection{Inference pipeline of DC$^2$}
The inference pipeline is detailed in Algorithm~\ref{alg:training}. The cropping function $\mathcal{F}_{crop}$ can be easily implemented using the Python Pillow module (PIL). Specifically, given an image $v$, we start by determining whether the width or height of the image $v$ is less than or equal to the resolution $S$ defined by the pretrained visual encoder. If so, it generates and returns a text description $T_l$ and $O_l$ for current recursive layer $l$. If not, the image is split into patches, and similarity between these patches is analyzed using hierarchical clustering ($HC$). The algorithm then recursively processes these patches, merging similar ones and updating the set of objects and text description. After processing, it refines the results, store relevant objects and coordinate in visual memory $\mathcal{M}$, and uses $\mathcal{M}$ to generate a final response based on a question $Q$.

\begin{algorithm}[H]
\small
\caption{Inference Pipeline of \textbf{DC$^2$}}
\label{alg:dc}
\label{alg:training}
\begin{algorithmic}[1]
\STATE \textbf{Input}: current recursive layer $l$; image $v$; the resolution $\mathcal{S}$ defined by pretrained visual encoder; hierarchical clustering $HC(\cdot)$; the threshold $\theta$ for forming flat clusters; visual memory $\mathcal{M}$; Retrieve $R$; Question $Q$; Multimodal LLM $MLLM(\cdot)$; \\
\STATE \textbf{Function DC$^2$($v_l$)}
\STATE \quad \textbf{if} $v_l.width \leq \mathcal{S}$ \textbf{or} $v_l.height \leq \mathcal{S}$ \textbf{then}
\STATE \quad\quad Generate description and objects, \ie $T_l, O_l \gets \mathcal{F}_{leaf}(v_l)$
\STATE \quad\quad \textbf{return} $T_l, O_l$
\STATE \quad \textbf{end if}
\STATE \quad Crop image $v_l$ into four patches, \ie $\{\overline{v}^i\}_{i=1}^4 \gets \mathcal{F}_{crop}(v_l)$
\STATE \quad Initialize an empty set $O \gets \{\}$
\STATE \quad Initialize an empty set $T \gets \{\}$
\STATE \quad Calculate similarity, \ie $[C_1,...,C_k] \gets HC(\{\overline{v}^i\}_{i=1}^4,\theta)$
\STATE \quad \textbf{for} $i=1,...,k$ \textbf{do}
\STATE \quad \quad Merge similar patches, \ie $v_{l+1}^i \gets \frac{1}{|C_i|}\sum_{\overline{v}\in C_i} \overline{v}$ 
\STATE \quad \quad Recursive processing, \ie $O_{l+1}, T_{l+1} \gets $ \textbf{DC$^2$($v_{l+1}^i$)}
\STATE \quad \quad Update $O$ with $O_{l+1}$, \ie $O \gets O \cup O_{l+1}$
\STATE \quad \quad Update $T$ with $T_{l+1}$, \ie $T \gets T \cup T_{l+1}$
\STATE \quad \textbf{end for}
\STATE \quad Conquer for non-leaf node, \ie $T_l, O_l \gets \mathcal{F}_{non-leaf}(v_l,T)$
\STATE \quad Filter out uncertainty objects, \ie $\widehat{O}_l \gets O_l \cap O$
\STATE \quad Store existing objects $\widehat{O}_l$ and coordinate $(x,y,w,h)$ in $\mathcal{M}$
\STATE \quad \textbf{return} $T_l, O_l$
\STATE \textbf{End Function}
\STATE
\STATE Store information in the visual memory $\mathcal{M}$, \ie \textbf{DC$^2(v)$}
\STATE Retrieve image patch $v_s$, \ie $v_s \gets R(M,Q)$
\STATE Generate text description $T$, \ie $T \gets MLLM(Q,v_s)$
\STATE Generate final response, \ie $Response \gets MLLM(Q+T,v)$
\STATE \textbf{Output}: $Response$
\end{algorithmic}
\end{algorithm}

\section{C. More Experiment Results}
\label{sec:experiment}
\subsection{Experiment Settings}
All experiments in this study are conducted within the same codebase modified from VLMEvalKit~\cite{duan2024vlmevalkit}. We conduct experiments on a PC with Intel(R) Xeon(R) Platinum 8358P CPU at 2.60GHz, and $8\times$ NVIDIA Tesla A100 80GB. To ensure the robustness of our experiments, we conduct each experiments five times. To clarify, the small temperature value of 0.2 was chosen to prevent large variances in the final results caused by significant differences in text descriptions. We set $\theta=0.1$ and $\alpha=0.3$ for all experiments unless otherwise stated.

\subsection{Fully Experiment Results on \itbf{HR-Bench}}
As shown in Table~\ref{table:hr_bench}, the SOTA MLLMs achieve accuracy rate of only 71.0\% and 62.8\% on \itbf{HR-Bench} 4K \& 8K, respectively. This is significantly lower compared to the accuracy rates of 82.0\% and 86.8\% achieved by humans. We note that the performance gap between open-source and commercial MLLMs in high-resolution (HR) image perception is minimal. The leading open-source MLLM, InternVL-2-llama-3-76B~\cite{chen2023internvl}, even outperforms commercial MLLMs~\cite{reid2024gemini,achiam2023gpt} on \itbf{HR-Bench 4K}. Furthermore, our \textbf{DC$^2$} consistently enhances performance on both \itbf{HR-Bench 4K} and \itbf{HR-Bench 8K}.

\paragraph{MLLMs exhibit relative weakness in fine-grained cross-instance perception of HR images.}
To further explore the ability of MLLMs with HR images, we calculate the average accuracy of 28 MLLMs on \itbf{FSP} and \itbf{FCP} respectively. The accuracy of \itbf{FSP} is 42.4\%, and the accuracy of \itbf{FCP} is 38.8\%. We observe that with cropping-based methods (\eg InternVL-v1.5), an increase in the number of image patches results in a decline in \itbf{FCP} performance.

\paragraph{MLLMs exhibit limitations in analyzing tasks involving high-density charts.} 
To explore the ability of existing MLLM to understand charts in HR images, we design a series of bar chars with different densities (\eg the sales data for 30, 50, and 200 items). The questions involve simple calculations (\eg summation and averaging). We observe that the best performance is 50\% accuracy, achieved by GPT4o. The performance of other MLLMs is nearly equivalent to random guessing.
\begin{table*}[thbp]
  \centering
  \begin{tabular}{@{}lcccccc@{}}
  \toprule
                 \multirow{2}{*}{\textbf{Method}}     & \multicolumn{3}{c}{\itbf{HR-Bench 4K}} & \multicolumn{3}{c}{\itbf{HR-Bench 8K}} \\ \cmidrule(lr){2-4} \cmidrule(lr){5-7}
                      & \itbf{FSP}    & \itbf{FCP}    & \itbf{Avg.}    & \itbf{FSP}    & \itbf{FCP}    & \itbf{Avg.}    \\ \hline
  \textbf{Human}               &   \textbf{90.0}     &    \textbf{74.0}    &    \textbf{82.0}    &   \textbf{94.0}     &   \textbf{79.5}     &   \textbf{86.8}     \\
  Random Guess        &    25.0    &   25.0     &  25.0      &   25.0     &  25.0      &   25.0     \\ \hline
  \multicolumn{7}{c}{\textit{Open-source MLLMs}} \\ \hline
  \textbf{InternVL-2-llama3-76B}~\cite{chen2023internvl} & \textbf{82.0} & \textbf{60.0} & \textbf{71.0} & 69.0  & \textbf{53.8}  & \textbf{61.4}  \\
  InternVL-1.5-26B~\cite{chen2024far}        & 69.5   & 51.8  & 60.6  & \textbf{69.3}  & 48.5   & 57.9  \\
  Xcomposer2-4kHD-7B~\cite{internlmxcomposer2_4khd} & 63.8 & 51.8 & 57.8 &  55.3  & 47.3   & 51.3  \\
  LLaVA-1.6-34B~\cite{liu2024llavanext}       & 55.3  & 50.5   & 52.9  & 44.5   & 50.3  & 47.4  \\
  LLaVA-HR-X-13B~\cite{luo2024feast}      & 61.3  & 46.0     & 53.6  & 49.5   & 44.3  & 46.9  \\
  LLaVA-HR-X-7B~\cite{luo2024feast}       & 57.8  & 46.3  & 52.0     & 42.0     & 41.3  & 41.6  \\
  CogVLM-Chat-17B~\cite{wang2023cogvlm}         & 49.5   & 41.5   & 45.5   & 42.5   & 39.8  & 41.1  \\
  LLaVA-1.6-7B~\cite{liu2024llavanext}        & 49.0     & 46.8   & 47.9   & 37.2   & 44.2   & 40.8   \\
  Phi3-Vision-4.2B~\cite{abdin2024phi}         & 54.3  & 42.0     & 48.1  & 43.3  & 37.8  & 40.5   \\
  Yi-VL-34B~\cite{young2024yi}           & 46.0     & 42.8  & 44.4  & 39.5   & 38.5   & 39.0     \\
  Yi-VL-6B~\cite{young2024yi}            & 42.8  & 42.5   & 42.6  & 38.5   & 39.3  & 38.9  \\
  LLaVA-1.6-13B~\cite{liu2024llavanext}       & 49.8  & 41.3  & 45.5   & 38.0     & 38.3  & 38.1  \\
  InternLM-Xcomposer2-7B~\cite{internlmxcomposer2} & 45.5   & 46.5   & 46.0     & 36.0     & 39.8  & 37.9  \\
  LLaVA-v1.5-13B~\cite{liu2024improved}      & 45.2   & 41.3   & 43.3   & 37.5   & 38.0     & 37.8   \\
  SEAL-7B~\cite{wu2024v}                &    47.0    &     29.3   &   38.1 &   42.5     &   28.8     &   35.6       \\
  InternLM-Xcomposer-7B~\cite{zhang2023internlmxcomposer}           & 37.3   & 41.3   & 39.3   & 34.5   & 35.8   & 35.1   \\
  MPLUG2-7B~\cite{Xu2023mPLUG2AM}              & 38.0     & 35.8  & 36.9  & 33.8  & 33.8  & 33.8  \\
  Deepseek-VL-7B~\cite{lu2024deepseekvl}         & 36.8  & 34.3  & 35.5   & 33.8  & 33.0     & 33.4  \\
  LLaVA-v1.5-7B~\cite{liu2024improved}       & 38.5   & 33.8   & 36.1   & 33.0     & 31.3   & 32.1   \\
  idefics-9B~\cite{laurencon2023obelics}             & 36.3  & 25.0     & 30.6  & 34.8  & 22.8  & 28.8  \\
  Qwen-VL-7B~\cite{Qwen-VL}           & 30.5   & 33.0     & 31.8   & 28.8   & 28.0     & 28.4   \\
  fuyu-8B~\cite{fuyu-8b}             & 25.5   & 26.0     & 25.8  & 24.8  & 27.5   & 26.1  \\
  minigptv2-7B~\cite{chen2023minigptv2}            & 25.8  & 25.3  & 25.5   & 26.0     & 26.3  & 26.1  \\
  VisualGLM-6B~\cite{du2022glm}        & 22.5   & 18.5   & 20.5   & 19.5   & 18.5   & 19.0     \\
  \hline
  \multicolumn{7}{c}{\textit{Commercial chatbot systems}} \\ \hline
  \textbf{Gemini 1.5 Flash}~\cite{reid2024gemini}        & \textbf{76.8}     & \textbf{56.8}     & \textbf{66.8}     & \textbf{69.2}  & \textbf{56.7}   & \textbf{62.8}  \\
  GPT4o~\cite{achiam2023gpt}              & 70.0     & 48.0     & 59.0     & 62.0     & 49.0     & 55.5   \\
  QWen-VL-max~\cite{Qwen-VL}               & 65.0     & 52.0     & 58.5     & 54.0     & 51.0     & 52.5   \\
  QWen-VL-plus~\cite{Qwen-VL}               & 65.0     & 41.0     & 53.0     & 52.0     & 41.0     & 46.5   \\
  \hline
  \multicolumn{7}{c}{\textit{w/ DC$^2$}} \\ \hline
  \textbf{InternVL-1.5-26B}       &  \textbf{73.3}  &  \textbf{53.5} &  \textbf{63.4}  &   \textbf{75.0}  &  \textbf{47.5}  &  \textbf{61.3} \\
  \tablerowcolor \quad $\Delta (\uparrow)$ & +3.8 & +1.7 & +2.8 & +5.7 & +1.0  & +3.4   \\ \hdashline
  LLaVA-v1.6-7B     &  53.0  & 47.0  & 50.0   &   40.5  & 45.0   & 42.3  \\
  \tablerowcolor \quad $\Delta (\uparrow)$ & +4.0 & +0.2 & +2.1 & +3.3 & +0.8  & +2.1   \\ \hdashline
  LLaVA-v1.5-13B     & 52.0   & 51.0  & 51.5   &  40.0   & 41.0   & 40.5  \\
  \tablerowcolor \quad $\Delta (\uparrow)$ & +6.8 & +9.7 & +8.2 & +2.5 & +3.0  &  +2.7  \\ \hdashline
  Yi-VL-6B       &  44.0  &  44.0 &  44.0  &  39.0   & 41.0   &  40.0 \\
  \tablerowcolor \quad $\Delta (\uparrow)$ & +1.2 & +1.5 & +1.4 & +0.5 & +1.7  &  +1.1  \\
   \bottomrule
  \end{tabular}
  \caption{Results of different models on the \itbf{HR-Bench}. The best performance in each task is in-bold. }
  \label{table:hr_bench}
  \end{table*}

\subsection{Effect of $\alpha$}
\begin{figure}[hbpt]
  \begin{center}
  \includegraphics[width=1.\linewidth]{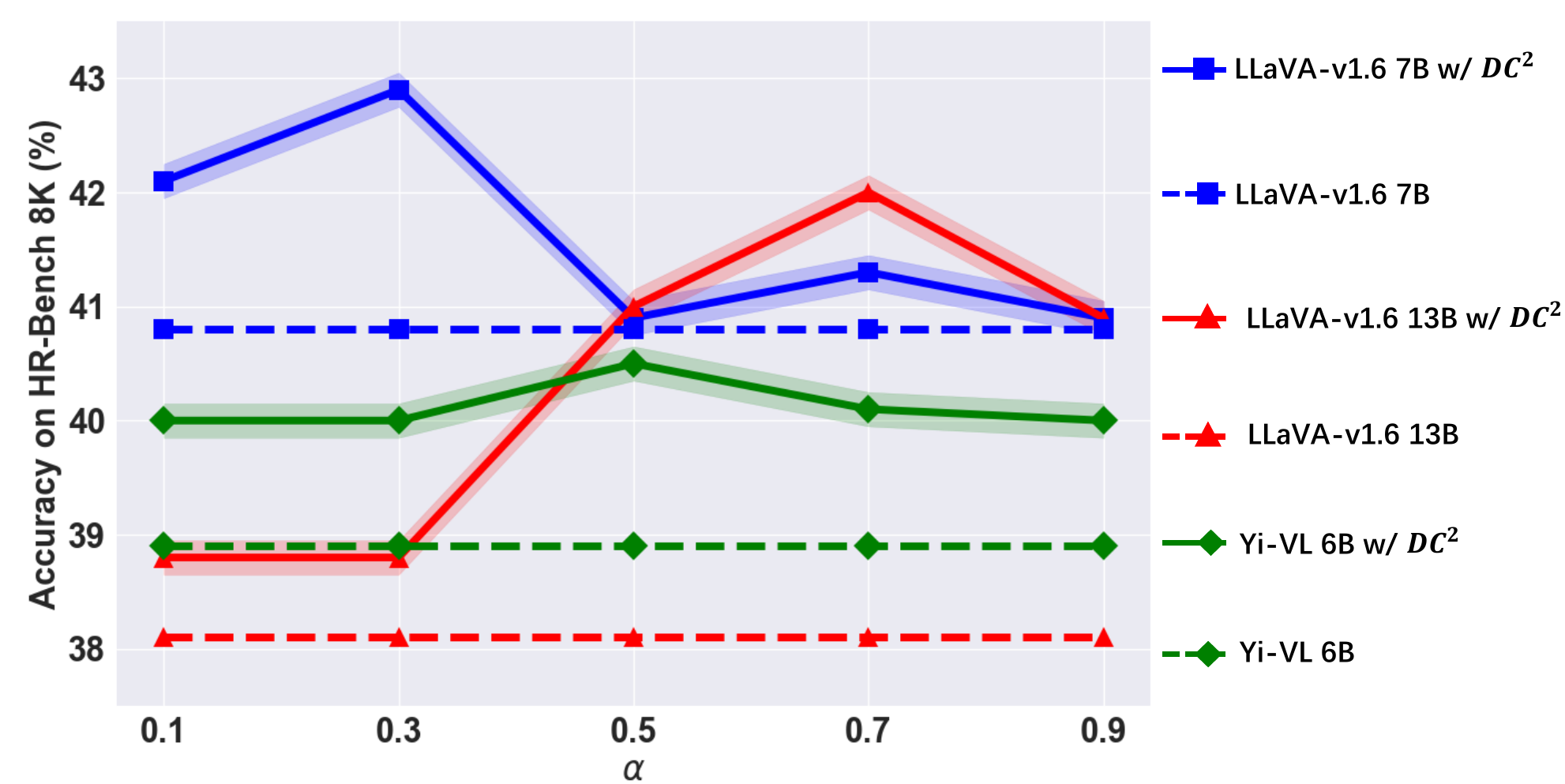}
  \end{center}
  \caption{Impact of retriever threshold $\alpha$, illustrating how the accuracy changes when varying $\alpha$.}
  \label{fig:alpha}
\end{figure}

In the inference stage, we use a textual retriever to retrieve related objects with confidence levels exceeding $\alpha$ from visual memory $\mathcal{M}$. To systematically study the impact of $\alpha$, we search for different configurations of $\alpha$. As shown in Figure~\ref{fig:alpha}, we find that \textbf{our DC$^2$ can bring consistent improvements across various values of $\alpha$.}

\subsection{Scaling DC$^2$}
\begin{figure}[hbpt]
    \begin{center}
    \includegraphics[width=1.\linewidth]{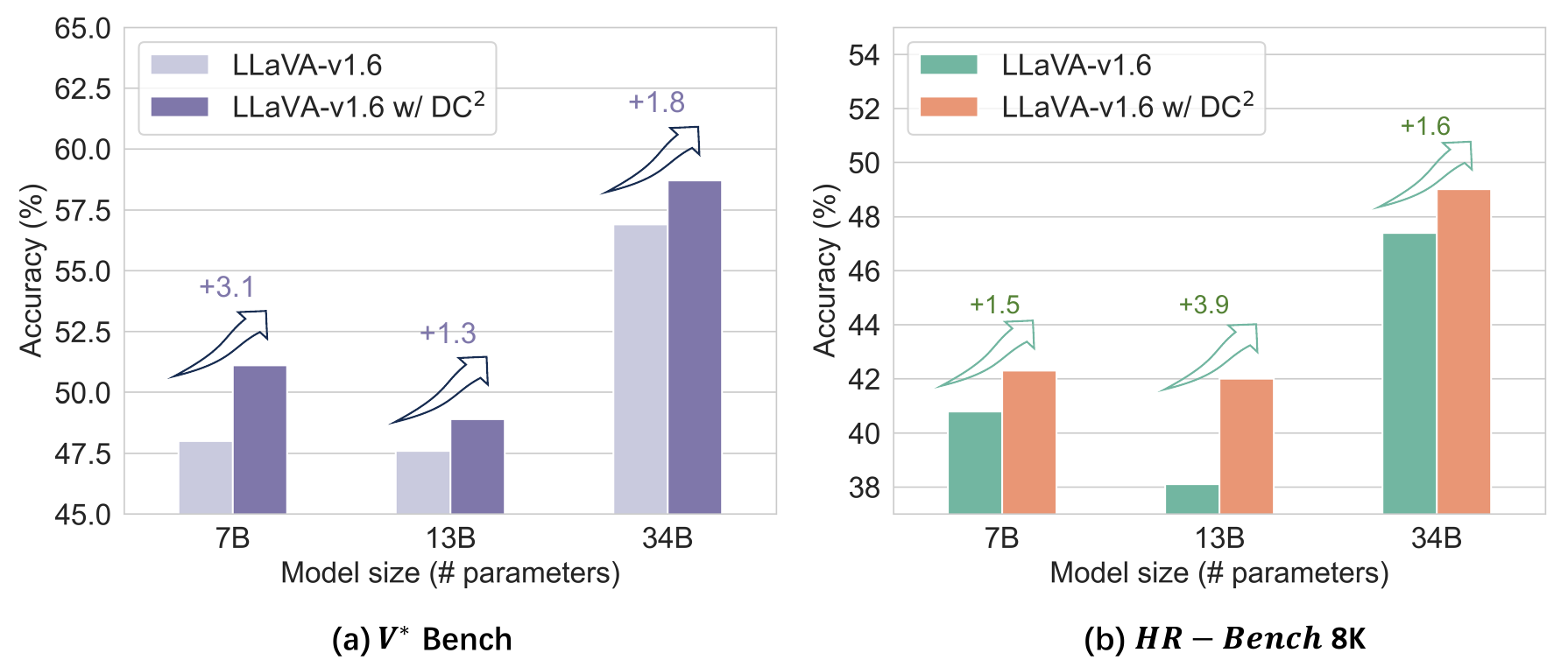}
    \end{center}
    \caption{Performance of scaling DC$^2$ with different model sizes. (a) \itbf{V$^*$} Bench and (b) \itbf{HR-Bench 8K}.}
    \label{fig:model_scale}
\end{figure}

We conduct experiments by scaling the model size to observe any potential effects when operating at a larger scale. As illustrated in Figure~\ref{fig:model_scale}, the results demonstrate that \textbf{the performance increases as the MLLMs size increases.}

\subsection{Effect of Prompts on Generating Accurate Text Descriptions}
\begin{figure}[hbpt]
  \begin{center}
  \includegraphics[width=1.\linewidth]{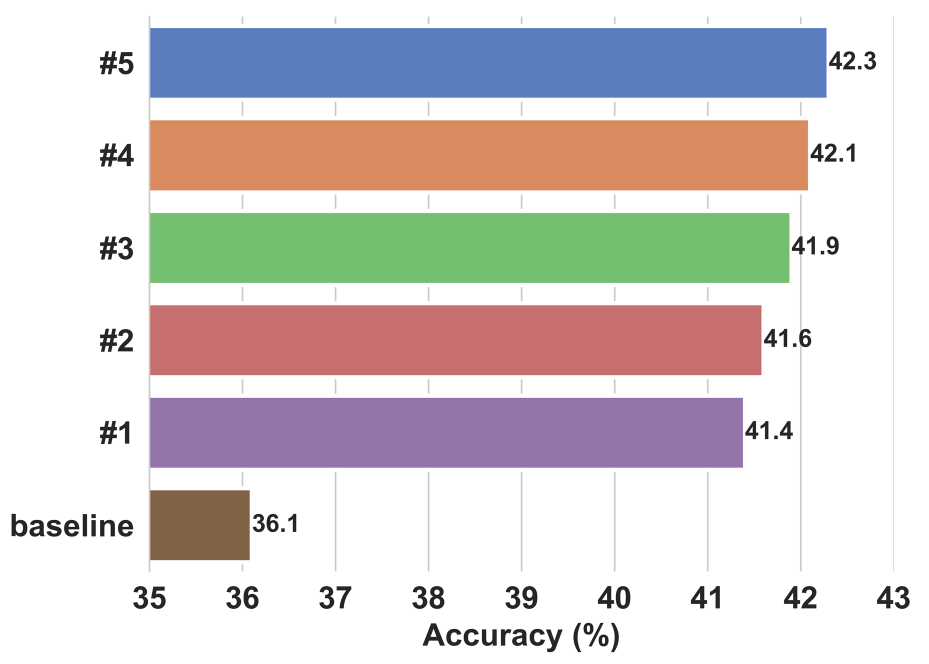}
  \end{center}
  \caption{Effect of different prompts on \itbf{HR-Bench 4K}. The baseline is LLaVA-v1.5 7B.}
  \label{fig:effect_prompt}
\end{figure}

During the conquer stage, we leverage MLLM to generate text descriptions for image patches. To assess the impact of various prompts on the generated captions, we conduct experiments on \itbf{HR-Bench 4K} using LLaVA-v1.5 7B~\cite{liu2024improved} with our \textbf{DC$^2$} across five different prompts (refer to Table~\ref{table:effect_prompts}). As illustrated in Figure~\ref{fig:effect_prompt}, we find that 1) the performance variations among these five prompts on \itbf{HR-Bench 4K} are minimal, and 2) even the simplest prompt (\ie \#1) can lead to a substantial improvement. Moreover, the text descriptions generated using prompt \#1 are significantly shorter than those from the other prompts, resulting in faster inference speed. Considering efficiency, we select prompt \#1 as the default for generating text descriptions.

\begin{table}[thpb]
\begin{tabularx}{\linewidth}{cX}
\toprule
\textbf{ID}  & \textbf{Prompt}     \\ \hline
\multirow{1}{*}[-0.5ex]{\#1} & Please describe this image. \\
\midrule
\multirow{1}{*}[-6ex]{\#2} & Describe the entire scene in the image, starting with the environment and setting. Include details of the background, foreground, and any significant objects or people. Mention any actions or interactions taking place, as well as the overall mood or atmosphere conveyed by the image. \\
\midrule
\multirow{1}{*}[-6ex]{\#3} & Identify and describe the key objects or subjects in the image. Specify their locations relative to the background and foreground. Highlight any actions, interactions, or significant details that draw attention, and explain how these elements contribute to the image's overall theme or narrative. \\
\midrule
\multirow{1}{*}[-6ex]{\#4} & Detail the environment depicted in the image, including weather, time of day, and any natural or artificial lighting. Describe how these factors influence the mood and tone of the image. Mention any significant objects or people present, and how they interact with the environment. \\
\midrule
\multirow{1}{*}[-6ex]{\#5} & Describe the characters or subjects in the image, focusing on their expressions, body language, and interactions. Explain how these elements convey emotions or relationships. Include details of the setting and any relevant objects that enhance the understanding of the scene. \\
\bottomrule
\end{tabularx}
\caption{Examples of the prompt. \#1 is a concise prompt provided manually, directly asking the MLLM to generate a text description of the image. \#2 to \#5 are prompts provided by ChatGPT 4o, containing more detailed task instructions.}
\label{table:effect_prompts}
\end{table}

\section{D. Case Study}
\label{sec:case_study}
\subsection{Examples in \itbf{HR-Bench}}
\begin{figure*}[hbpt]
  \begin{center}
  \includegraphics[width=0.8\linewidth]{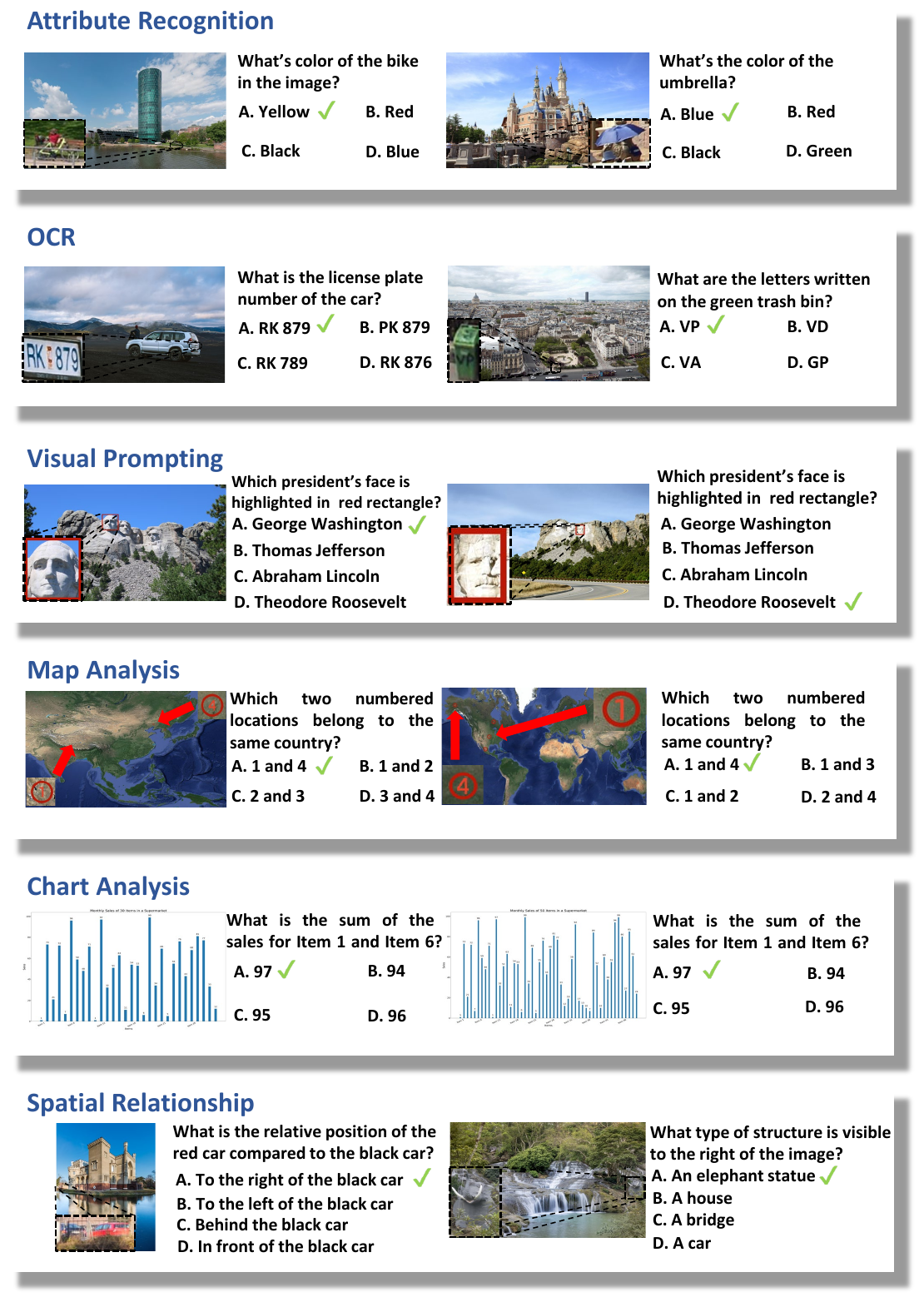}
  \end{center}
  \caption{Examples of data in \itbf{HR-Bench}.}
  \label{fig:case_data}
\end{figure*}

As shown in Figure~\ref{fig:case_data}, we visualize examples of six different types of tasks in \itbf{HR-Bench 8K}. We display the key image region used to answer the corresponding question in the corner of the image.

\subsection{Qualitative Examples of \itbf{HR-Bench 8K}}
\begin{figure*}[hbpt]
  \begin{center}
  \includegraphics[width=0.8\linewidth]{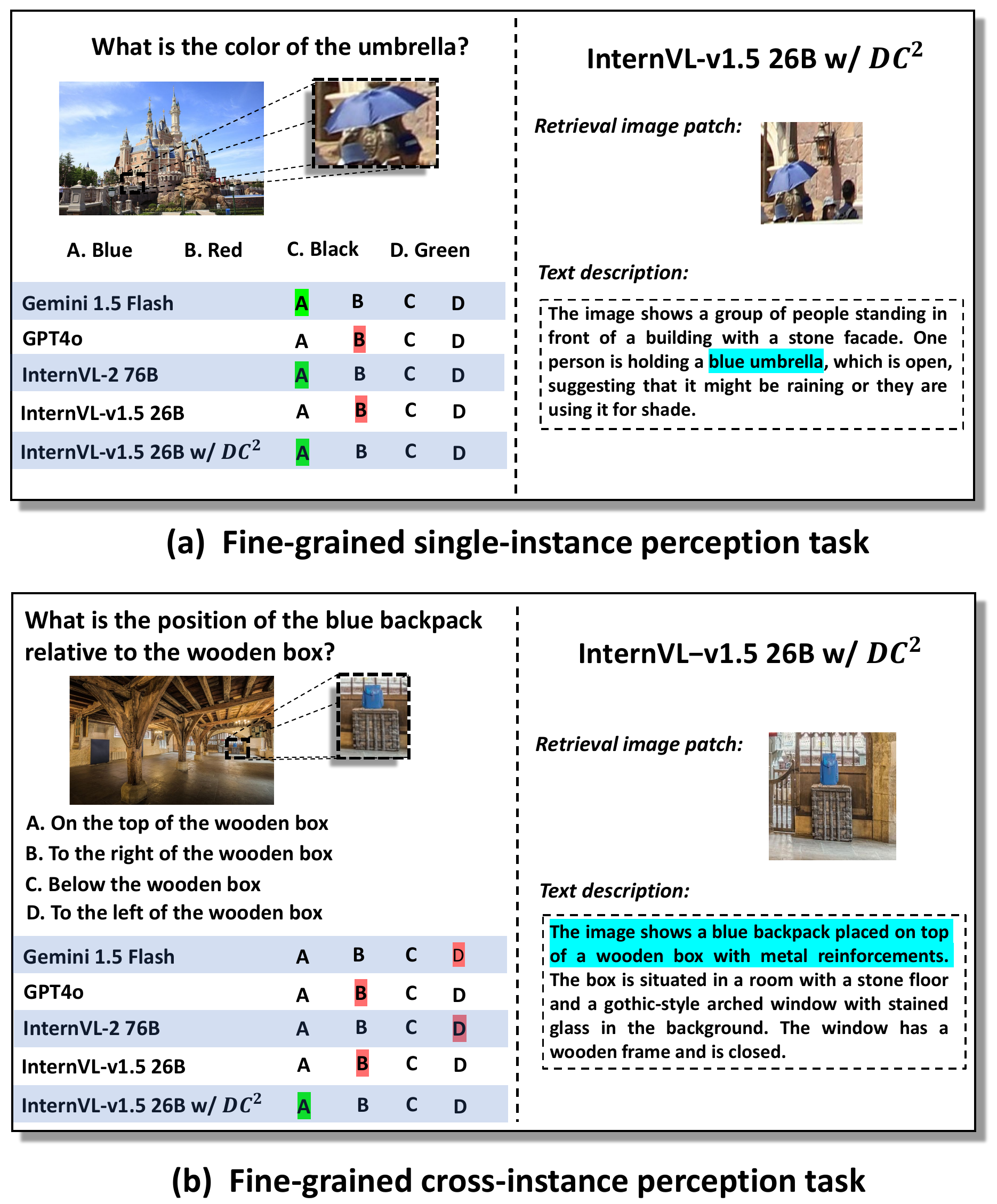}
  \end{center}
  \caption{Qualitative examples of \itbf{HR-Bench 8K}. Incorrect answers are shaded in \colorbox{incorrect}{red}. Correct answers are shaded in \colorbox{correct}{green}. Auxiliary information for answering related questions in text descriptions are shaded in \colorbox{keyword}{blue}.}
  \label{fig:case_method}
\end{figure*}

As showin in Figure~\ref{fig:case_method}, we visualize qualitative examples of four MLLMs (the top 2 open-source and commercial models) on two sub-tasks (\ie \itbf{FSP} and \itbf{FCP}) of \itbf{HR-Bench 8K}. We observe that even the current SOTA MLLMs~\cite{reid2024gemini,achiam2023gpt,chen2023internvl} cannot accurately perceive fine-grained objects in HR images. However, our \textbf{DC$^2$}, as a training-free framework, can be seamlessly integrated into existing MLLMs. By providing accurate text descriptions of image patches, it helps MLLMs better perceive HR images. 

\end{document}